\documentclass[journal]{IEEEtran}
\usepackage[T1]{fontenc}
\usepackage{tgtermes}
\usepackage{amsmath,amssymb}
\usepackage{graphicx}
\usepackage{booktabs}
\usepackage{array}
\usepackage{xcolor}
\usepackage{microtype}
\usepackage{cite}
\usepackage[hidelinks]{hyperref}

\newcommand{\QwenTTwoAccClean}{0.99}

\newcommand{\QwenTTwoBlank}{0.26}
\newcommand{\QwenTTwoShuf}{0.29}

\newcommand{\QwenTTwoFogAccMild}{0.97}

\newcommand{\QwenTTwoFogAurocMild}{0.98}
\newcommand{\QwenTTwoFogAccMid}{0.74}

\newcommand{\QwenTTwoFogAurocMid}{0.89}
\newcommand{\QwenTTwoFogAccDeep}{0.38}

\newcommand{\QwenTTwoFogAbsDeep}{0.85}
\newcommand{\QwenTTwoFogAbsTwoDeep}{0.53}

\newcommand{\QwenTTwoFogAccDeeper}{0.25}

\newcommand{\QwenTTwoFogIntWorst}{0.81}
\newcommand{\QwenTTwoFogVerWorst}{0.90}

\newcommand{\QwenTTwoFogEceWorst}{0.56}
\newcommand{\QwenTTwoFogAurocWorst}{0.53}

\newcommand{\QwenTTwoFogVacc}{12.4}
\newcommand{\QwenTTwoFogWInt}{0.80}

\newcommand{\QwenTTwoFogWAbsTwo}{0.18}
\newcommand{\QwenTTwoFogWAbsTwoLo}{0.15}
\newcommand{\QwenTTwoFogWAbsTwoHi}{0.21}

\newcommand{\QwenTTwoUwAbsTwoDeep}{0.30}

\newcommand{\QwenTTwoUwAbsTwoWorst}{0.24}
\newcommand{\QwenTTwoUwEceWorst}{0.56}

\newcommand{\QwenTTwoUwVacc}{22.1}
\newcommand{\QwenTTwoUwWInt}{1.05}

\newcommand{\QwenTTwoUwWAbsTwo}{1.05}

\newcommand{\QwenTThreeBlank}{0.50}

\newcommand{\QwenTThreeFogAbsDeep}{0.01}

\newcommand{\QwenTThreeFogIntWorst}{1.00}

\newcommand{\InternTTwoAccClean}{0.98}

\newcommand{\InternTTwoBlank}{0.24}
\newcommand{\InternTTwoShuf}{0.29}

\newcommand{\InternTTwoFogAccMild}{0.95}

\newcommand{\InternTTwoFogAurocMild}{0.96}
\newcommand{\InternTTwoFogAccMid}{0.67}

\newcommand{\InternTTwoFogAurocMid}{0.83}
\newcommand{\InternTTwoFogAccDeep}{0.31}

\newcommand{\InternTTwoFogAbsDeep}{0.97}

\newcommand{\InternTTwoFogAccDeeper}{0.24}

\newcommand{\InternTTwoFogIntWorst}{0.68}
\newcommand{\InternTTwoFogVerWorst}{0.93}

\newcommand{\InternTTwoFogEceWorst}{0.44}
\newcommand{\InternTTwoFogAurocWorst}{0.51}

\newcommand{\InternTTwoFogVacc}{14.1}
\newcommand{\InternTTwoFogWInt}{0.86}

\newcommand{\InternTTwoFogWAbsTwo}{0.05}

\newcommand{\InternTTwoUwEceWorst}{0.44}

\newcommand{\InternTTwoUwVacc}{29.5}
\newcommand{\InternTTwoUwWInt}{1.18}

\newcommand{\InternTThreeBlank}{0.50}

\newcommand{\InternTThreeFogAbsDeep}{0.01}

\newcommand{\InternTThreeFogIntWorst}{0.92}

\newcommand{\SmolTTwoAccClean}{0.85}

\newcommand{\SmolTTwoAbsClean}{0.96}
\newcommand{\SmolTTwoAbsTwoClean}{0.48}
\newcommand{\SmolTTwoBlank}{0.25}
\newcommand{\SmolTTwoShuf}{0.25}

\newcommand{\SmolTTwoFogAccMild}{0.86}
\newcommand{\SmolTTwoFogIntMild}{0.70}

\newcommand{\SmolTTwoFogAbsTwoMild}{0.44}

\newcommand{\SmolTTwoFogAbsTwoWorst}{1.00}

\newcommand{\SmolTTwoFogVacc}{15.9}
\newcommand{\SmolTTwoFogWInt}{0.07}
\newcommand{\SmolTTwoFogWIntLo}{0.03}
\newcommand{\SmolTTwoFogWIntHi}{0.10}

\newcommand{\SmolTTwoUwVacc}{38.6}

\newcommand{\SmolTThreeBlank}{0.50}

\newcommand{\GemmaNImg}{600}

\newcommand{\GemmaTTwoBlank}{0.28}
\newcommand{\GemmaTTwoShuf}{0.27}

\newcommand{\GemmaTTwoFogAurocMild}{0.94}

\newcommand{\GemmaTTwoFogAurocMid}{0.86}

\newcommand{\GemmaTTwoFogAbsDeep}{0.92}

\newcommand{\GemmaTTwoFogIntWorst}{0.79}
\newcommand{\GemmaTTwoFogVerWorst}{1.00}

\newcommand{\GemmaTTwoFogEceWorst}{0.50}
\newcommand{\GemmaTTwoFogAurocWorst}{0.54}

\newcommand{\GemmaTTwoFogVacc}{13.3}
\newcommand{\GemmaTTwoFogWInt}{0.83}

\newcommand{\GemmaTTwoFogWAbs}{-0.06}

\newcommand{\GemmaTTwoFogWAbsTwo}{-0.03}

\newcommand{\GemmaTTwoUwEceWorst}{0.52}

\newcommand{\GemmaTTwoUwVacc}{31.0}
\newcommand{\GemmaTTwoUwWInt}{1.20}

\newcommand{\GemmaTThreeBlank}{0.50}

\newcommand{\GemmaTThreeFogAbsDeep}{0.10}

\newcommand{\GemmaTThreeFogIntWorst}{1.00}

\newcommand{\QwenFourNImg}{400}

\newcommand{\QwenFourTTwoBlank}{0.30}
\newcommand{\QwenFourTTwoShuf}{0.29}

\newcommand{\QwenFourTTwoFogAbsDeep}{0.95}

\newcommand{\QwenFourTTwoFogIntWorst}{0.84}
\newcommand{\QwenFourTTwoFogVerWorst}{0.61}

\newcommand{\QwenFourTTwoFogVacc}{12.1}
\newcommand{\QwenFourTTwoFogWInt}{0.79}

\newcommand{\QwenFourTTwoFogWAbs}{-0.21}

\newcommand{\QwenFourTTwoFogWAbsTwo}{-0.06}

\newcommand{\QwenFourTTwoUwVacc}{20.5}

\newcommand{\QwenFourTTwoUwWVer}{0.75}
\newcommand{\QwenFourTTwoUwWVerLo}{0.38}
\newcommand{\QwenFourTTwoUwWVerHi}{1.05}

\newcommand{\QwenFourTThreeFogAbsDeep}{0.09}

\newcommand{\NConfModelsWord}{four}
\newcommand{\NLagFactorFog}{6.2}
\newcommand{\NLagWMinFog}{0.79}
\newcommand{\AbsWMinFog}{-0.21}
\newcommand{\AbsWMaxFog}{-0.06}

\newcommand{\AbsTwoWMaxFog}{0.18}
\newcommand{\NLagFactorUw}{5.6}

\newcommand{\AbsWMinUw}{-0.26}
\newcommand{\AbsWMaxUw}{0.02}

\newcommand{\ConfIntWorstMin}{0.68}
\newcommand{\ConfIntWorstMax}{0.84}
\newcommand{\ConfVerWorstMin}{0.61}
\newcommand{\ConfVerWorstMax}{1.00}

\newcommand{\VaccFogMin}{12.1}
\newcommand{\VaccFogMax}{15.9}
\newcommand{\VaccUwMin}{20.5}
\newcommand{\VaccUwMax}{38.6}
\newcommand{\NomVeqOne}{110}
\newcommand{\NomSpreadOne}{6.2}
\newcommand{\NomStdVeqOne}{128}
\newcommand{\NomVeqTwo}{82}

\newcommand{\NomStdVeqTwo}{114}
\newcommand{\NomVeqThree}{67}

\newcommand{\NomStdVeqThree}{123}
\newcommand{\NomVeqFour}{65}

\newcommand{\NomStdVeqFour}{141}
\newcommand{\NomVeqFive}{55}
\newcommand{\NomSpreadFive}{8.4}
\newcommand{\NomStdVeqFive}{152}
\newcommand{\NomMonoFrac}{52}
\newcommand{\QwenTTwoNomMin}{0.97}
\newcommand{\QwenTTwoNomMax}{0.98}

\newcommand{\InternTTwoNomMin}{0.96}
\newcommand{\InternTTwoNomMax}{0.97}

\newcommand{\SmolTTwoNomMin}{0.85}
\newcommand{\SmolTTwoNomMax}{0.89}

\newcommand{\GemmaTTwoNomMin}{0.94}
\newcommand{\GemmaTTwoNomMax}{0.96}

\newcommand{\QwenFourTTwoNomMin}{0.97}
\newcommand{\QwenFourTTwoNomMax}{0.99}

\newcommand{\NomMaxDevPP}{5}
\newcommand{\BlankMaxDev}{0.05}

\newcommand{\MonoFogAccRng}{55.4}

\newcommand{\MonoNomAccRng}{2.2}

\newcommand{\QwenRelHiN}{4112}
\newcommand{\QwenRelHiAcc}{0.45}

\newcommand{\QwenTThreeYesFogWorst}{0.00}

\newcommand{\QwenTThreeNoFogWorst}{1.00}

\newcommand{\InternRelHiN}{1713}
\newcommand{\InternRelHiAcc}{0.50}

\newcommand{\InternTThreeYesFogWorst}{0.01}

\newcommand{\InternTThreeNoFogWorst}{0.99}

\newcommand{\GemmaRelHiN}{2167}
\newcommand{\GemmaRelHiAcc}{0.46}

\newcommand{\GemmaTThreeYesFogWorst}{0.00}

\newcommand{\GemmaTThreeNoFogWorst}{1.00}

\newcommand{\NRowsTotal}{258{,}000}

\newcommand{\ConfigHash}{812b518e9400}
\newcommand{\ItemsHash}{619cbf46561d}
\newcommand{\DepthMedianOfMedians}{13.8}

\newcommand{\SensQwenTTwoFogEightVTwo}{0.76}

\newcommand{\SensQwenTTwoFogTwoVOne}{0.90}

\newcommand{\SensQwenTTwoFogTwoVTwo}{0.73}

\newcommand{\SensQwenTTwoFogTwoVThree}{0.96}

\newcommand{\SensQwenTTwoFogTwoAcc}{0.22}

\newcommand{\SensInternTTwoFogTwoVOne}{0.94}

\newcommand{\SensInternTTwoFogTwoVTwo}{0.99}

\newcommand{\SensInternTTwoFogTwoVThree}{0.90}

\newcommand{\SensInternTTwoFogTwoAcc}{0.26}

\newcommand{\Vacc}{V_{\mathrm{acc}}}
\newcommand{\Vaware}{V_{\mathrm{aware}}}
\newcommand{\Ttwo}{\textsc{t2}}
\newcommand{\Tthree}{\textsc{t3}}

\hypersetup{pdftitle={How Murky Before a VLM Knows It Cannot See? Accuracy, Confidence and Abstention on a Physical Visibility Axis},
  pdfauthor={Mochamad Rizal Fauzan, Irgi Surya},
  pdfkeywords={vision-language models; visibility; fog; underwater; calibration; abstention; robustness}}

\begin{document}

\title{How Murky Before a VLM Knows It Cannot See?\\
Accuracy, Confidence and Abstention\\
on a Physical Visibility Axis}

\author{%
\begin{tabular}[t]{@{}c@{}}
Mochamad Rizal Fauzan\\[3pt]
{\small National Taipei University of Technology}\\
{\small Taipei, Taiwan}\\
{\small t114998415@ntut.edu.tw}
\end{tabular}
\hspace{2.4cm}
\begin{tabular}[t]{@{}c@{}}
Irgi Surya\\[3pt]
{\small National Taipei University of Technology}\\
{\small Taipei, Taiwan}\\
{\small t115999404@ntut.edu.tw}
\end{tabular}}

\markboth{Preprint, October 2026}{Fauzan and Surya: How Murky Before a VLM Knows It Cannot See?}

\maketitle

\begin{abstract}
Robustness studies of vision-language models (VLMs) grade degradation on nominal severity scales (``level 1--5'') that carry no physical units, and they rarely ask whether a model's confidence follows the loss of visibility.
We measure accuracy, internal (token-probability) confidence, verbalized confidence and an explicit abstain option on a \emph{visibility axis in metres}: fog follows the Koschmieder model, an underwater arm scales the absorption spectrum of pure water, both are applied with a monocular metric depth map, and the visibility range $V$ is swept over eight log-spaced levels from 250\,m to 1.95\,m.
On 1000 COCO scenes and five small open VLMs (2--5\,B parameters), accuracy on a boxed-object identification task falls from near its clean value to chance within a narrow band (half-way visibility $\Vacc$ of \VaccFogMin--\VaccFogMax\,m in fog and \VaccUwMin--\VaccUwMax\,m under water) while, for the \NConfModelsWord\ models that are confident on clean images, the confidence stays high: at $V=1.95$\,m in fog their accuracy is at chance but their mean internal confidence is \ConfIntWorstMin--\ConfIntWorstMax\ and their verbalized confidence \ConfVerWorstMin--\ConfVerWorstMax, a blind-spot width $W=\log_{10}(\Vacc/\Vaware)$ of at least \NLagWMinFog\ decades, i.e.\ a factor of \NLagFactorFog\ in visibility.
The remaining model is under-confident throughout and its internal confidence follows accuracy.
An explicit abstain option follows accuracy when it names the reason ($W$ from \AbsWMinFog\ to \AbsWMaxFog\ in fog) but, for one model, lags with a generic ``I don't know'' ($W$ up to \AbsTwoWMaxFog).
ImageNet-C fog severities 1--5 correspond to only \NomVeqFive--\NomVeqOne\,m of physical visibility, outside the failure band, and leave the accuracy of the models within \NomMaxDevPP\ percentage points of its clean value.
Blank-image and shuffled-image controls, a prior-matched answer set and a 200-image pilot that preceded the study are reported.
Findings are descriptive: one draw of 1000 images, depth of assumed scale, and models of 2--5\,B parameters.
\end{abstract}


\section{Introduction}
\label{sec:intro}

\IEEEPARstart{C}{ameras} that look through fog, haze or water see a scene whose radiance is attenuated and veiled by the medium in a way that grows with distance \cite{narasimhan2002,narasimhan2003,jaffe1990,akkaynak2018}.
The same image-formation model underlies haze removal \cite{he2009dcp} and the estimation of visibility distance from a camera image \cite{hautiere2006}.
A vision-language model (VLM) asked a question about such an image has to do two things: answer correctly while the scene is still visible, and stop answering with confidence once it is not.
The second ability is what a downstream system needs from a visual question answering component that is meant to hand control back to a human or to a slower sensor.

Evaluation practice does not yet address this directly.
Broad VLM benchmarks report accuracy on clean images \cite{liu2024mmbench}, and hallucination benchmarks ask whether answers are grounded in the image \cite{li2023pope,guan2024hallusion}; to our reading, none of them grades the image on a physical visibility scale.
Robustness benchmarks apply corruptions at ordinal severities taken from the ImageNet-C convention \cite{hendrycks2019,michaelis2019}; VLM-specific benchmarks follow the same pattern \cite{saxena2026,sui2025,qiu2022mm}, and the reliability of VLMs under corrupted inputs has begun to be examined for driving \cite{xie2025drivebench}.
A severity ``3'' of one library is not comparable with ``3'' of another, and neither maps to a quantity an engineer can relate to a sensor or a site.
Image quality is known to hurt deep classifiers \cite{dodge2016}, and post-hoc calibration is known to fall short under dataset shift \cite{ovadia2019}.
Work on VLM confidence has begun to measure stated versus internal confidence of small VLMs under photographic degradations \cite{ferdous2026,ferdous2026b} and confidence--correctness alignment under corruptions \cite{sui2025}, but, to our reading, not along an axis in physical units and not with an explicit way for the model to decline.

Declining to answer has a long history as the reject option of classifiers \cite{chow1970,geifman2017}.
For visual question answering it has been studied as selective prediction \cite{whitehead2022} and through questions that cannot be answered, notably those that arise from the often poor photographs of blind users \cite{gurari2018vizwiz}; benchmarks of unanswerable or unsolvable questions followed \cite{guo2024unkvqa,miyai2024upd,he2024tubench}.
To our reading, they do not grade the visual evidence on a physical scale.

This paper asks a simple question: \emph{at what visibility range does a VLM start to be wrong, and at what visibility range does it start to say so?}
We call the difference between the two the \emph{blind spot}.
To make the question measurable we use degradations whose strength is a physical quantity: the Koschmieder fog model \cite{narasimhan2002}, parameterized by the meteorological visibility $V$ in metres, and an underwater arm built from the absorption spectrum of pure water \cite{popefry1997}.
Both are applied with a monocular metric depth map, and $V$ is swept over eight log-spaced levels from 250\,m to 1.95\,m.
At every level we read out four things from five small open VLMs: the answer (accuracy), the internal confidence from token probabilities, a verbalized confidence, and the choice of an explicit abstain option.

\paragraph*{Research questions}
\textbf{RQ1} Over what range of $V$ does accuracy collapse?
\textbf{RQ2} Do internal and verbalized confidence follow the collapse, or lag behind it?
\textbf{RQ3} Does an explicit abstain option expose what the stated confidence hides, and how robust is it to wording?
\textbf{RQ4} Do nominal ImageNet-C fog severities cover the failure band?

\paragraph*{Contributions}
(i) A protocol that places accuracy, internal confidence, verbalized confidence and abstention on one visibility axis in metres, with a scale-free summary, the blind-spot width $W=\log_{10}(\Vacc/\Vaware)$, whose definition was frozen before any model was run.
(ii) A measurement on 1000 COCO scenes, five VLMs, two physical degradation arms and two tasks, with blank-image and shuffled-image controls and a prior-matched answer set.
(iii) Evidence that, for the models tested, confidence lags accuracy by at least a factor of \NLagFactorFog\ in visibility in fog, that an explicit abstain option follows accuracy only when its wording names the reason, and that ImageNet-C fog severities 1--5 are equivalent to \NomVeqFive--\NomVeqOne\,m, well above the failure band.
(iv) A set of design diagnostics for confidence studies under degradation (an existence-style question that is ill-posed under occlusion, a distractor prior, a box marker) that a 200-image pilot revealed and this study corrects.
We make no claim of priority; related work is discussed in Section~\ref{sec:related}.

\paragraph*{Scope}
Five models of 2--5\,B parameters, synthetic degradations, an estimated depth scale and one draw of 1000 images; two of the models were run on a subset of the images (the first \GemmaNImg\ and the first \QwenFourNImg).
The numbers in the tables, figures and text come from the result files through generator scripts, not from hand entry.
Section~\ref{sec:related} positions the study; Section~\ref{sec:method} defines the axis, tasks, prompts, read-outs and metrics; Section~\ref{sec:results} reports results, the verification task, the controls and the complete sweep tables; Sections~\ref{sec:discussion} and~\ref{sec:limits} discuss implications and limits.

\section{Related Work}
\label{sec:related}

\paragraph*{Robustness benchmarks and image quality}
ImageNet-C introduced a suite of common corruptions at five ordinal severities \cite{hendrycks2019}, later extended to detection \cite{michaelis2019}; earlier, the accuracy of deep classifiers had been shown to be susceptible to quality distortions such as blur and noise \cite{dodge2016}.
For multimodal models, 12 open image-text models were found not to be robust to image and text perturbations, especially image perturbations \cite{qiu2022mm}; VLM-RobustBench evaluates models from four families under 49 augmentations and reports accuracy \cite{saxena2026}; and Bench-C studies confidence--correctness alignment of 13 VLMs under 19 corruptions with five ImageNet-C-style severities \cite{sui2025}.
DriveBench evaluates 12 VLMs on driving questions under clean, corrupted and text-only inputs and reports that they often answer from general knowledge or textual cues rather than from the image, especially when the visual input is degraded or missing \cite{xie2025drivebench}.
A causal taxonomy of degradations quantifies severity with image-quality measures and demonstrates it on detectors \cite{becker2026}.
SynDORBench varies viewing distance, lighting and pose in synthetic scenes and organises them into pixel-density regimes, again without confidence \cite{yee2026}.
Our axis differs in that severity is a physical quantity (visibility range in metres) and in that the read-outs include confidence and abstention.

\paragraph*{Confidence and calibration}
Modern networks are over-confident \cite{guo2017}, calibration is commonly summarised by binned calibration errors \cite{naeini2015}, the maximum softmax probability is a strong baseline for error detection \cite{hendrycks2017}, and selective classification trades coverage for risk \cite{chow1970,geifman2017}.
Under dataset shift, post-hoc calibration falls short \cite{ovadia2019}, and the calibration of recent image classifiers depends on architecture beyond model size \cite{minderer2021}.
For language models, token-level probabilities can be well calibrated on multiple-choice tasks \cite{kadavath2022}, models can be trained or prompted to state confidence in words \cite{lin2022words,tian2023}, and black-box elicitation strategies have been benchmarked for calibration and failure prediction \cite{xiong2024}.
For VLMs, verbalized uncertainty has been found overconfident for GPT-4V and Gemini Pro Vision \cite{groot2024}, and the verbalized confidence for object-centric queries has been calibrated by training on object regions perturbed with noise \cite{zhao2025csp}.
For small VLMs, stated confidence has been reported to stay nearly constant under degradation while token probability still detects errors \cite{ferdous2026}; in the Qwen2-VL family, scale improves the error-detection AUROC of the internal signal while the verbalized signal stays weak \cite{ferdous2026b}.
We use the same two read-outs and add a third, an explicit abstain option, along a physical axis.

\paragraph*{Abstention and unanswerable questions}
For visual question answering, abstention has been analysed through coverage and risk: using a model's softmax score as the abstention criterion allows answering only a small fraction of the questions at a low risk of error \cite{whitehead2022}.
VizWiz collected visual questions from blind photographers, whose images are often of poor quality and whose questions often cannot be answered \cite{gurari2018vizwiz}.
UNK-VQA builds unanswerable questions by perturbing the image or the question \cite{guo2024unkvqa}; MM-UPD asks large multimodal models to withhold an answer on unsolvable multiple-choice problems and finds that models that do well on standard benchmarks struggle \cite{miyai2024upd}; TUBench uses questions that are unanswerable because the image lacks the information \cite{he2024tubench}.
MoHoBench tests honesty on unanswerable visual questions \cite{zhu2025mohobench}; MM-AQA creates unanswerable instances by changing the modality dependence and evidence sufficiency of answerable ones \cite{madhusudhan2026}; VB asks models to answer or abstain on visibility claims about photographs \cite{tripathi2026}; HalluPrism re-runs answers under visual degradation to diagnose halluc\-ination \cite{prakash2026}.
For language models, abstention has been surveyed along the query, the model and human values \cite{wen2025abstain}, and models can be tuned to say ``I don't know'' for questions outside their parametric knowledge \cite{zhang2024rtuning}.
These works indicate that VLMs rarely abstain by default.
They do not grade the visual evidence on a physical scale.

\paragraph*{Physical degradation models}
The Koschmieder law links meteorological visibility to the extinction coefficient and underlies fog synthesis for scene understanding \cite{narasimhan2002,sakaridis2018,li2019reside}; the same atmospheric scattering model is the basis of single-image dehazing \cite{he2009dcp} and of contrast restoration in bad weather \cite{narasimhan2003}, and visibility distance can itself be estimated from an onboard camera \cite{hautiere2006}.
Underwater, attenuation depends on wavelength and water type \cite{akkaynak2017,berman2021}; computer models of underwater imaging go back at least to \cite{jaffe1990}, and the common image-formation model needs care at the level of wide-band coefficients \cite{akkaynak2018,akkaynak2019}.
The absorption of pure water was measured by Pope and Fry \cite{popefry1997}, and the optical properties of the clearest natural waters were tabulated over 200--800\,nm by Smith and Baker \cite{smithbaker1981}.
Recent studies of degraded underwater perception still grade turbidity on ordinal levels \cite{uwrobots2026}, while FogDrive grades fog by visibility distance for 3D perception rather than for VLMs \cite{panwar2026}.

\begin{table*}[!t]
\caption{Positioning against the closest works found. Entries are taken from the cited abstracts, and for \cite{sui2025} and \cite{ferdous2026} from their full texts; ``--'' means not part of that study as far as can be read from them. This table is a reading aid, not a claim of exhaustiveness.}
\label{tab:related}
\centering\footnotesize
\newcolumntype{L}[1]{>{\raggedright\arraybackslash}p{#1}}
\begin{tabular}{@{}L{3.0cm}L{3.5cm}L{3.2cm}L{2.6cm}L{2.2cm}L{2.0cm}@{}}
\toprule
Work & Degradation & Severity axis & Confidence & Abstain & Models \\
\midrule
ImageNet-C \cite{hendrycks2019} & common corruptions & ordinal 1--5 & -- & -- & classifiers \\
VLM-RobustBench \cite{saxena2026} & 49 augmentations & nominal & -- & -- & VLMs, 4 families \\
Bench-C \cite{sui2025} & 19 corruptions & ordinal 1--5 (ImageNet-C style) & logit-based alignment & -- & 13 VLMs \\
Ferdous \cite{ferdous2026} & 6 photographic degradations & 3 ordinal levels & stated and internal & -- (forced choice) & 2 VLMs \\
SynDORBench \cite{yee2026} & distance, lighting, pose & pixel-density regimes & -- & -- & 16 LVLMs \\
MM-AQA \cite{madhusudhan2026} & evidence manipulation & not graded & baselines & yes & 3 VLMs \\
VB \cite{tripathi2026} & none (real photographs) & -- & confidence & yes & VLMs \\
\textbf{This study} & Koschmieder fog, pure-water spectral shape & visibility range $V$ (m) & internal and verbalized & yes (2 wordings) & 5 small VLMs \\
\bottomrule
\end{tabular}
\end{table*}

\section{Method}
\label{sec:method}

\begin{figure*}[!t]
\centering
\includegraphics[width=\textwidth]{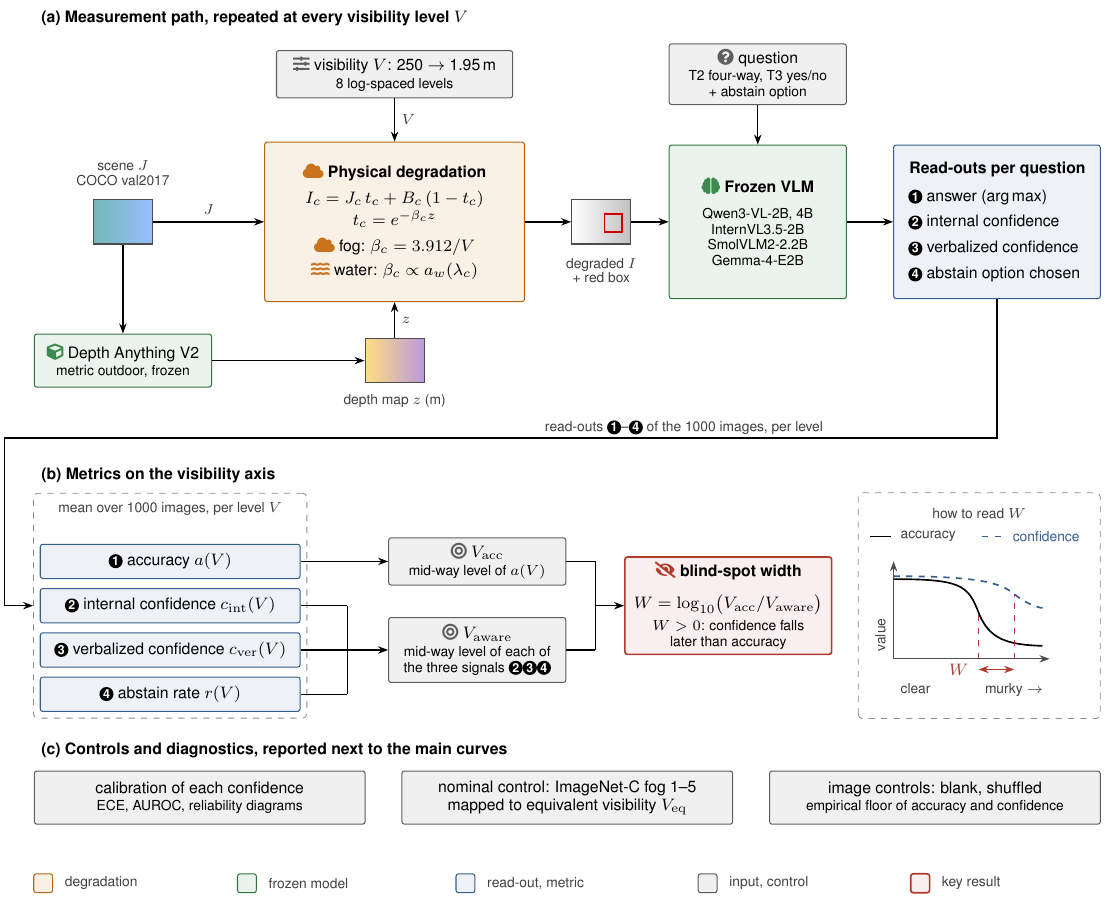}
\caption{System overview. (a) Measurement path: a COCO scene and its estimated metric depth are degraded by the image-formation model, Eq.~(\ref{eq:formation}), with the visibility $V$ swept over eight log-spaced levels; a red box is added to the degraded image and a question about the boxed object is posed to a frozen VLM, which yields an answer, an internal confidence, a verbalized confidence and an abstain choice (read-outs 1--4). (b) The four read-outs are averaged over the images at each level; $V_\mathrm{acc}$ and $V_\mathrm{aware}$ are the visibilities at which accuracy and each awareness signal reach their mid-way level, and the blind-spot width $W$ is their log ratio (inset: how to read $W$). (c) Controls and diagnostics: calibration of each confidence, ImageNet-C fog mapped to an equivalent visibility, and blank and shuffled images.}
\label{fig:arch}
\end{figure*}

Figure~\ref{fig:arch} summarises the pipeline: a scene and its metric depth are degraded by a physical operator at a given visibility range $V$; a question about the degraded image is posed to a frozen VLM; four read-outs are collected; and the metrics are computed on the visibility axis.

\subsection{Image formation and the visibility axis}
For a scene radiance $J_c$ ($c\in\{R,G,B\}$) at metric depth $z$, the observed intensity is
\begin{equation}
I_c = J_c\,t_c + B_c\,(1-t_c),\qquad t_c = e^{-\beta_c z},
\label{eq:formation}
\end{equation}
with transmission $t_c$, attenuation coefficient $\beta_c$ (m$^{-1}$) and veiling light $B_c$.
All operations are applied in linear RGB (sRGB decoded) and re-encoded to 8-bit sRGB.

\emph{Fog.} $\beta_R=\beta_G=\beta_B=\beta$ and $B_c=A=0.8$.
The meteorological visibility range is the distance at which a black target falls to a 2\,\% contrast threshold, $V=3.912/\beta$ \cite{narasimhan2002}, so $V$ is the single, physically meaningful strength parameter.

\emph{Underwater.} The attenuation of pure water is strongly wavelength dependent \cite{popefry1997}.
We set $\beta_c = s\,a_w(\lambda_c)$ with $\lambda_c=(620,540,460)$\,nm and $a_w=(0.2755,\,0.0474,\,0.00979)$\,m$^{-1}$ from the Pope--Fry measurements \cite{popefry1997}, and one common scale $s$.
The axis is the visibility range of the least attenuated channel, $V_\mathrm{best}=3.912/\min_c\beta_c$, and the veiling light is $B_c = 0.8\,\beta_c^{-1}/\max_k\beta_k^{-1}$, which is blue-dominated.
This arm captures the clear-ocean spectral shape only: absorption without scattering by particles, monochromatic channels instead of a sensor response, and no range dependence of the wide-band coefficients \cite{akkaynak2017,akkaynak2018}.
Because $V_\mathrm{best}$ and the fog $V$ are defined by different channels, absolute values are not comparable between arms; the scale-free $W$ below is.

\begin{figure*}[!t]
\centering
\includegraphics[width=\textwidth]{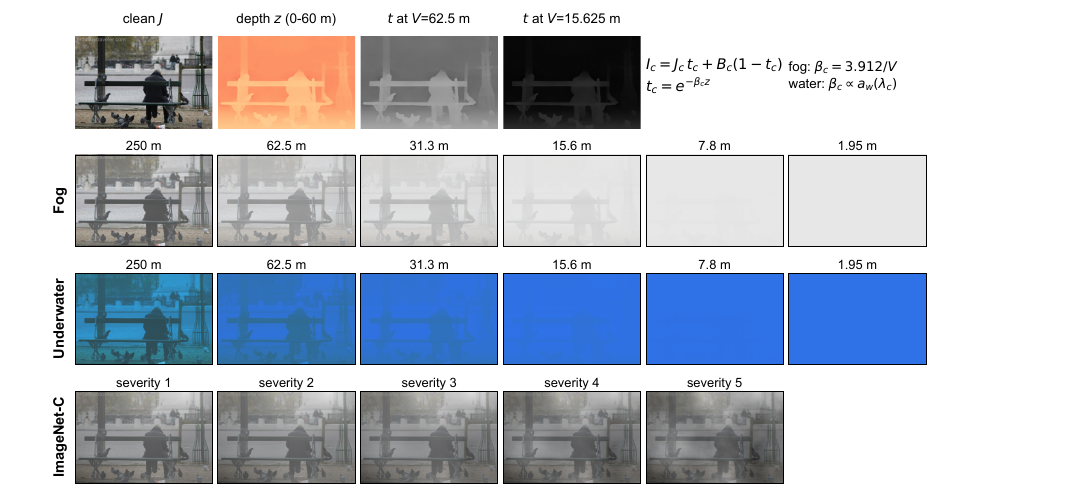}
\caption{One scene under the three degradation families. Top row: clean image, estimated metric depth, and the transmission maps $t$ for two fog levels. Fog and underwater rows: six of the eight sweep levels. ImageNet-C row: fog severities 1--5, which add a plasma-fractal haze that is independent of depth.}
\label{fig:montage}
\end{figure*}

\emph{Sweep.} $V\in\{250, 125, 62.5, 31.3, 15.6, 7.8, 3.9, 1.95\}$\,m (eight levels, factor 2), plus the clean image.
Figure~\ref{fig:montage} shows one scene at six of the levels.

\subsection{Depth and its assumed scale}
Depth comes from Depth Anything V2 \cite{yang2024dav2}, the successor of Depth Anything \cite{yang2024da} (metric outdoor model, small), clipped to $[0.5,80]$\,m; the median over images of the median scene depth is \DepthMedianOfMedians\,m.
It is estimated on the clean image, because monocular depth estimators are themselves sensitive to weather and lighting corruptions \cite{kong2023robodepth}.
The absolute scale of a monocular estimate is not verified for COCO.
Since only the product $\beta z=3.912\,z/V$ enters Eq.~(\ref{eq:formation}), a global error in depth by a factor $\kappa$ rescales every crossing visibility reported below by $\kappa$ and leaves $W$ unchanged.
Visibilities in metres are therefore conditional on the depth scale; $W$ is not.

\subsection{Data and tasks}
We draw 1000 images from COCO val2017 \cite{lin2014coco} (seed 1) among the 1146 that contain a vehicle or outdoor object and a boxable instance (area between 3\,\% and 50\,\% of the image, shorter side at least 48\,px), so that fog is a plausible condition.
One instance per image is the \emph{target}, chosen with probability inversely proportional to the frequency of its category among the candidates so that \emph{person} does not dominate.
Two tasks are posed on the target, which is marked by an 8-px red box drawn \emph{after} the degradation so that its position remains visible:
\begin{itemize}
\item \Ttwo\ (identification): ``What is the object inside the red box?'' with four options, the target category and three distractors; chance 0.25.
\item \Tthree\ (verification): ``Is the object inside the red box a(n) \emph{\textless category\textgreater}?'' asked once with the true category and once with a distractor, in the manner of POPE \cite{li2023pope}; chance 0.5.
\end{itemize}
Distractors are categories with no annotation in the image, drawn with probability proportional to the category's frequency among the targets divided by the fraction of images in which it is absent, so that a category is about as likely to appear as a distractor as to be the answer; a guesser that always picks the most frequent target category scores 0.29 on \Ttwo\ (chance 0.25).
This gives 3000 questions per condition.
Two image controls are added: a uniform grey image of the same size (blank, box kept) and a shuffled image (the question of image $i$ on the clean pixels of image $j=(i+37)\bmod 1000$).
The ImageNet-C \emph{fog} corruption at severities 1--5 \cite{hendrycks2019}, as implemented by the \texttt{imagecorruptions} package \cite{michaelis2019}, serves as the nominal control.
In total there are 24 conditions per model.

\subsection{Models and read-outs}
Five open VLMs are run frozen in bf16 with greedy decoding: Qwen3-VL-2B-Instruct and Qwen3-VL-4B-Instruct \cite{qwen3vl}, InternVL3.5-2B \cite{internvl35}, SmolVLM2-2.2B-Instruct \cite{smolvlm} and Gemma-4-E2B-it (Hugging Face model \texttt{google/gemma-4-E2B-it}).
Each image is resized to at most 448$^2$ pixels and presented as a single tile.
The assistant turn is prefilled with ``Answer:'' so that the first generated token is the answer.
Four read-outs are collected per question and condition:
\begin{enumerate}
\item \emph{Answer.} The option with the largest first-token probability.
\item \emph{Internal confidence.} $c_\mathrm{int}=\max_k p_k$, with $p_k$ the first-token probability of option $k$ renormalised over the options.
\item \emph{Verbalized confidence.} The integer after ``Confidence:'' in a 12-token continuation, divided by 100; unparsed outputs are kept as missing.
\item \emph{Abstain.} A second prompt adds an option ``I cannot tell, the image is too unclear to decide''; abstention is the event that this option has the largest first-token probability. A third prompt offers instead ``I don't know'' (abstain$_2$), because wording is known to matter (Section~\ref{sec:results}).
\end{enumerate}
The abstain option is offered as one more choice of a multiple-choice question, the setting in which unsolvable problems have been used to test whether a model withholds an answer \cite{miyai2024upd}; Table~\ref{tab:prompts} lists all prompt templates, which were fixed before the main run.

\begin{table}[!t]
\caption{Prompt templates. \texttt{\{obj\}} is a COCO category name and \texttt{\{o1\}}--\texttt{\{o4\}} are the four options; the assistant turn is prefilled with ``Answer:''.}
\label{tab:prompts}
\centering\footnotesize
\begin{tabular}{@{}p{0.2\columnwidth}p{0.71\columnwidth}@{}}
\toprule
Prompt & Text \\
\midrule
\Ttwo, main & What is the object inside the red box? A.~\texttt{\{o1\}} B.~\texttt{\{o2\}} C.~\texttt{\{o3\}} D.~\texttt{\{o4\}} Answer with the letter of the correct option. \\
\Tthree, main & Is the object inside the red box a(n) \texttt{\{obj\}}? Answer with Yes or No. \\
Confidence & After your answer, on a new line, state how confident you are that your answer is correct, as a number from 0 to 100, in the format `Confidence: N'. \\
Abstain, main & \Tthree: A.~Yes B.~No C.~I cannot tell, the image is too unclear to decide. \Ttwo: fifth option E.~I cannot tell, the image is too unclear to decide. \\
Abstain$_2$ & The same prompts with ``I don't know'' as the last option. \\
\bottomrule
\end{tabular}
\end{table}

\subsection{Metrics}
All quantities are image-level means with percentile bootstrap intervals over images \cite{efron1979} (1000 resamples for per-condition values, 2000 for crossing points).
\emph{Accuracy} is the fraction of correct answers; \emph{ECE} uses ten equal-width confidence bins \cite{naeini2015,guo2017}; \emph{AUROC} is the area under the ROC curve of the confidence used as a detector of correct answers \cite{fawcett2006}.

\emph{Crossing points.}
Let $a_0$ be a signal's value on the clean image and $a_\mathrm{floor}$ its reference floor (chance for accuracy and confidence, one for the abstain rate).
Scanning from clear to murky, the crossing visibility is where the signal first reaches the mid-way value $(a_0+a_\mathrm{floor})/2$, with linear interpolation in $\log V$ between adjacent levels.
It is written $\Vacc$ for accuracy and $\Vaware$ for each awareness signal (internal confidence, verbalized confidence, abstain rate).
If the signal never reaches the mid-way value within the sweep, the crossing is censored at 1.95\,m and the resulting $W$ is a lower bound.

\emph{Blind-spot width.}
\begin{equation}
W=\log_{10}\!\big(\Vacc/\Vaware\big).
\label{eq:W}
\end{equation}
$W>0$ means that the awareness signal falls later, in murkier conditions, than accuracy; $W=0$ means they fall together.
Because Eq.~(\ref{eq:W}) is a ratio of two visibilities from the same sweep, it is invariant to the global depth scale.

\emph{Nominal equivalence.}
For each image and ImageNet-C severity we find the physical-fog visibility $V_\mathrm{eq}$ that retains the same image contrast, measured as the regression gain of the degraded luma on the clean luma.
The statistic frozen before the pilot, the ratio of luma standard deviations, proved non-monotone in severity because the additive haze adds variance; the gain is therefore the primary statistic and the ratio is reported as a check.

\emph{Monotonicity.}
For each signal series we count adjacent level pairs that move in the wrong direction (accuracy or confidence rising, abstention falling when going murkier), as a raw fraction and as a fraction exceeding two standard errors of the paired image-level difference.

\subsection{Experimental configuration}
Table~\ref{tab:config} lists the fixed settings of the main run.
Experiments ran on a single RTX 5080 (16\,GB) with PyTorch 2.11 and Transformers 5.19.

\begin{table}[!t]
\caption{Fixed settings of the main run.}
\label{tab:config}
\centering\footnotesize
\begin{tabular}{@{}ll@{}}
\toprule
Item & Value \\
\midrule
Images & 1000, COCO val2017, seed 1 \\
Questions per condition & 3000 (1000 \Ttwo, 2000 \Tthree) \\
Conditions per model & 24 \\
Visibility levels & 250 $\to$ 1.953125\,m, factor 2 \\
Fog veiling light & $A=0.8$ (linear RGB) \\
Water $a_w(620,540,460)$ & 0.2755, 0.0474, 0.00979\,m$^{-1}$ \\
Depth model & Depth Anything V2 metric outdoor (small) \\
Depth clip & 0.5--80\,m \\
Red box width & 8\,px \\
Decoding & greedy, bf16, $\leq$12 new tokens \\
Image size & $\leq$448$^2$ pixels, one tile \\
Bootstrap & 1000 / 2000 image resamples \\
ECE bins & 10 \\
\bottomrule
\end{tabular}
\end{table}

\subsection{Pre-registration and relation to the pilot}
This study follows a 200-image pilot with three models, whose controls revealed task-design confounds.
The design changes below were frozen in a configuration file (SHA-256 prefix \ConfigHash) before any stage-1 model was run; the 1000 selected items have hash prefix \ItemsHash.
(i) The existence task of the pilot, which was ill-posed under occlusion, is replaced by \Tthree, which always concerns the boxed object.
(ii) Distractors are prior-matched, because the pilot's blank-image accuracy was below chance.
(iii) The box is thicker (8 px instead of 3 px).
(iv) The sweep reaches 1.95\,m so that confidence signals can cross their mid-way level.
(v) A second abstain wording is collected in the same run.
Post-hoc additions are labelled as such: the noise-aware monotonicity count, the blank-floor variant of $\Vacc$, the reliability diagrams and the gain-based nominal equivalence.

\section{Results}
\label{sec:results}

Unless stated otherwise, values are for the identification task \Ttwo, whose floor is well defined (the blank-image accuracy is within \BlankMaxDev\ of chance for every model, Table~\ref{tab:controls}); the verification task \Tthree\ is reported in Section~\ref{sec:diag}, and the complete sweep tables are in Section~\ref{sec:levels}.
Intervals are 95\,\% bootstrap intervals over images.
Models are abbreviated Qwen-2B, Intern-2B, Smol-2.2B, Gemma-E2B and Qwen-4B.

\subsection{Accuracy collapses within a narrow visibility band (RQ1)}
\begin{figure*}[!t]
\centering
\includegraphics[width=\textwidth]{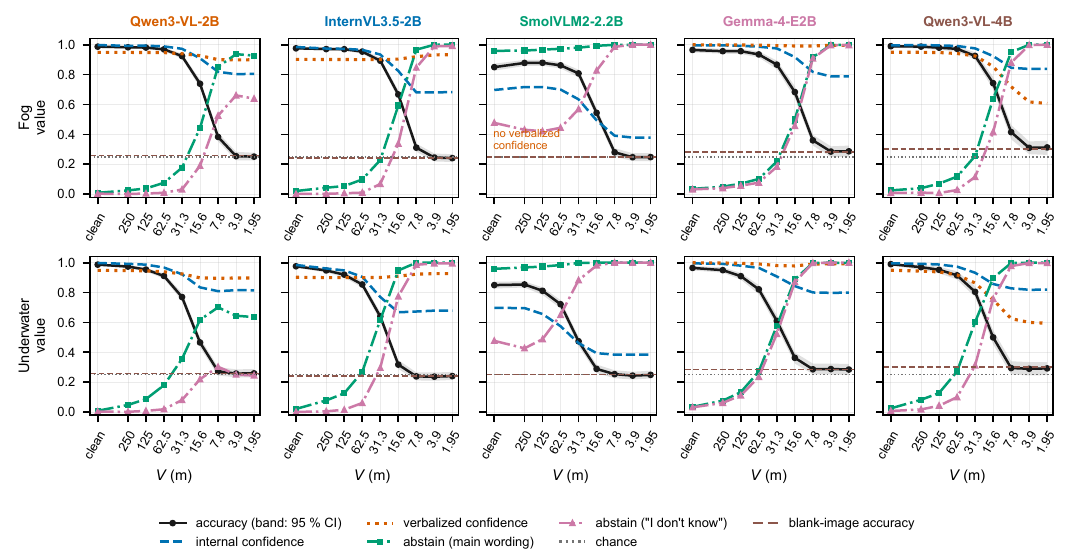}
\caption{Identification task (\Ttwo): accuracy and the awareness read-outs against the visibility range $V$ (log scale, murkier to the right) for fog (top) and underwater (bottom). Accuracy falls to chance within a factor of about ten in $V$ while the internal and verbalized confidence of the three confident models stay high; the abstain option rises with the loss of accuracy when it names the reason (main wording) and, for some models, much less when it is generic (``I don't know''). Smol-2.2B produced no parseable verbalized confidence and abstains on most clean questions with the main wording.}
\label{fig:curves}
\end{figure*}
Figure~\ref{fig:curves} shows that in fog the accuracy of every model stays near its clean value down to $V=62.5$\,m and then falls to its blank-image floor between 15.6 and 3.9\,m.
For Qwen-2B it is \QwenTTwoAccClean\ on clean images, \QwenTTwoFogAccMild\ at 62.5\,m, \QwenTTwoFogAccMid\ at 15.6\,m, \QwenTTwoFogAccDeep\ at 7.8\,m and \QwenTTwoFogAccDeeper\ at 3.9\,m (chance 0.25); for Intern-2B \InternTTwoAccClean, \InternTTwoFogAccMild, \InternTTwoFogAccMid, \InternTTwoFogAccDeep\ and \InternTTwoFogAccDeeper.
The half-way visibility is $\Vacc=\QwenTTwoFogVacc$, \InternTTwoFogVacc, \SmolTTwoFogVacc, \GemmaTTwoFogVacc\ and \QwenFourTTwoFogVacc\,m in fog and \QwenTTwoUwVacc, \InternTTwoUwVacc, \SmolTTwoUwVacc, \GemmaTTwoUwVacc\ and \QwenFourTTwoUwVacc\,m under water ($V_\mathrm{best}$) for Qwen-2B, Intern-2B, Smol-2.2B, Gemma-E2B and Qwen-4B.
At the end of the sweep accuracy equals the accuracy on a blank image, so no visual information is left.

\subsection{Confidence lags accuracy (RQ2)}
\begin{figure*}[!t]
\centering
\includegraphics[width=\textwidth]{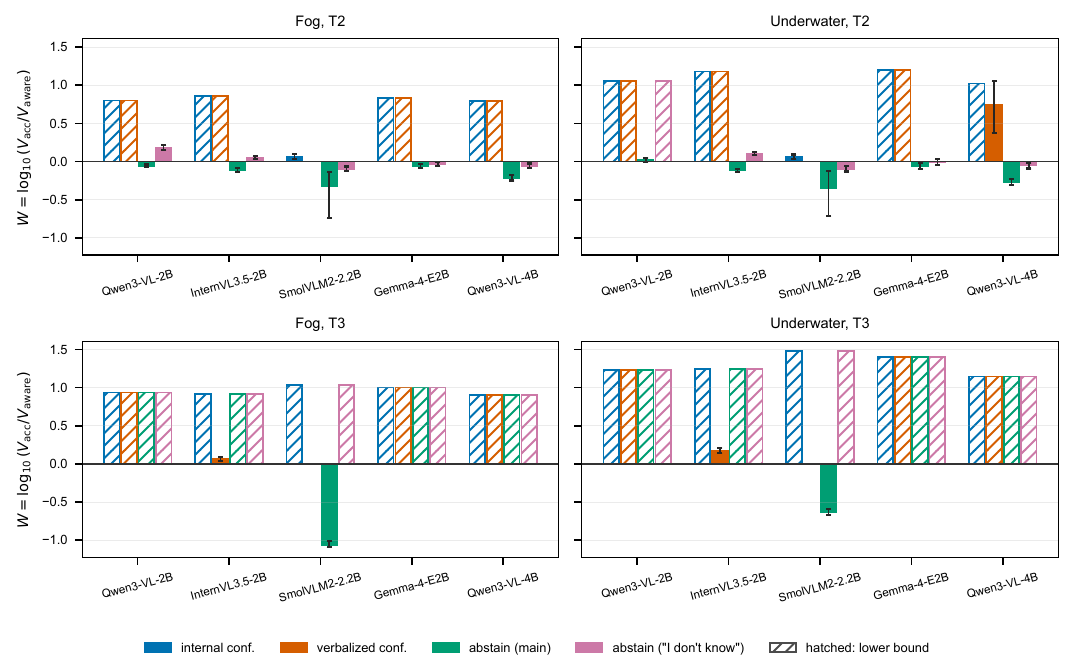}
\caption{Blind-spot width $W=\log_{10}(\Vacc/\Vaware)$ for the identification (top) and verification (bottom) tasks. Hatched bars are lower bounds: the awareness signal never reached its mid-way level within the sweep. Error bars are 95\,\% bootstrap intervals over images. Bars are absent where no verbalized confidence could be parsed.}
\label{fig:W}
\end{figure*}

\begin{table*}[!t]
\caption{Two murky levels ($V=7.8$\,m and $1.95$\,m): accuracy, mean internal confidence, mean verbalized confidence and abstain rates (main wording; abstain$_2$ is ``I don't know'') for both tasks and both arms. Chance is 0.25 for \Ttwo\ and 0.5 for \Tthree. ``--'': verbalized confidence could not be parsed.}
\label{tab:worst}
\centering\footnotesize
\resizebox{\textwidth}{!}{\begin{tabular}{@{}lllccccc|ccccc@{}}
\toprule
 & & & \multicolumn{5}{c}{$V=7.8$\,m} & \multicolumn{5}{c}{$V=1.95$\,m} \\
Model & Task & Arm & Acc. & Int. & Verb. & Abst. & Abst.$_2$ & Acc. & Int. & Verb. & Abst. & Abst.$_2$ \\
\midrule
Qwen3-VL-2B & T2 & fog & 0.38 & 0.82 & 0.90 & 0.85 & 0.53 & 0.25 & 0.81 & 0.90 & 0.93 & 0.64 \\
Qwen3-VL-2B & T2 & water & 0.27 & 0.81 & 0.90 & 0.70 & 0.30 & 0.26 & 0.82 & 0.90 & 0.64 & 0.24 \\
Qwen3-VL-2B & T3 & fog & 0.55 & 0.99 & 0.98 & 0.01 & 0.00 & 0.50 & 1.00 & 1.00 & 0.00 & 0.00 \\
Qwen3-VL-2B & T3 & water & 0.50 & 1.00 & 0.99 & 0.00 & 0.00 & 0.50 & 1.00 & 1.00 & 0.00 & 0.00 \\
\midrule
InternVL3.5-2B & T2 & fog & 0.31 & 0.68 & 0.92 & 0.97 & 0.85 & 0.24 & 0.68 & 0.93 & 1.00 & 0.99 \\
InternVL3.5-2B & T2 & water & 0.24 & 0.67 & 0.92 & 1.00 & 0.99 & 0.24 & 0.68 & 0.93 & 1.00 & 0.99 \\
InternVL3.5-2B & T3 & fog & 0.54 & 0.89 & 0.51 & 0.01 & 0.00 & 0.50 & 0.92 & 0.40 & 0.00 & 0.00 \\
InternVL3.5-2B & T3 & water & 0.50 & 0.87 & 0.49 & 0.00 & 0.00 & 0.50 & 0.88 & 0.51 & 0.00 & 0.00 \\
\midrule
SmolVLM2-2.2B & T2 & fog & 0.28 & 0.39 & -- & 1.00 & 0.99 & 0.25 & 0.38 & -- & 1.00 & 1.00 \\
SmolVLM2-2.2B & T2 & water & 0.25 & 0.38 & -- & 1.00 & 1.00 & 0.25 & 0.38 & -- & 1.00 & 1.00 \\
SmolVLM2-2.2B & T3 & fog & 0.50 & 0.77 & -- & 1.00 & 0.04 & 0.50 & 0.79 & -- & 1.00 & 0.08 \\
SmolVLM2-2.2B & T3 & water & 0.50 & 0.78 & -- & 1.00 & 0.04 & 0.50 & 0.78 & -- & 1.00 & 0.06 \\
\midrule
Gemma-4-E2B & T2 & fog & 0.36 & 0.82 & 0.99 & 0.92 & 0.91 & 0.29 & 0.79 & 1.00 & 1.00 & 1.00 \\
Gemma-4-E2B & T2 & water & 0.29 & 0.80 & 0.99 & 1.00 & 1.00 & 0.28 & 0.80 & 0.99 & 1.00 & 1.00 \\
Gemma-4-E2B & T3 & fog & 0.51 & 1.00 & 1.00 & 0.10 & 0.00 & 0.50 & 1.00 & 1.00 & 0.01 & 0.00 \\
Gemma-4-E2B & T3 & water & 0.50 & 1.00 & 1.00 & 0.03 & 0.00 & 0.50 & 1.00 & 1.00 & 0.01 & 0.00 \\
\midrule
Qwen3-VL-4B & T2 & fog & 0.41 & 0.85 & 0.72 & 0.95 & 0.88 & 0.31 & 0.84 & 0.61 & 1.00 & 1.00 \\
Qwen3-VL-4B & T2 & water & 0.29 & 0.83 & 0.63 & 1.00 & 0.98 & 0.29 & 0.82 & 0.59 & 1.00 & 1.00 \\
Qwen3-VL-4B & T3 & fog & 0.54 & 0.99 & 0.95 & 0.09 & 0.00 & 0.50 & 1.00 & 1.00 & 0.00 & 0.00 \\
Qwen3-VL-4B & T3 & water & 0.50 & 1.00 & 0.98 & 0.01 & 0.00 & 0.50 & 1.00 & 1.00 & 0.00 & 0.00 \\
\bottomrule
\end{tabular}
}
\end{table*}
At $V=1.95$\,m (Table~\ref{tab:worst}) the accuracy of every model is at its floor in fog, yet the mean internal confidence is \QwenTTwoFogIntWorst, \InternTTwoFogIntWorst\ and \GemmaTTwoFogIntWorst\ for Qwen-2B, Intern-2B and Gemma-E2B (\QwenFourTTwoFogIntWorst\ for Qwen-4B), and the mean verbalized confidence is \QwenTTwoFogVerWorst, \InternTTwoFogVerWorst\ and \GemmaTTwoFogVerWorst\ (\QwenFourTTwoFogVerWorst).
The ECE of the internal confidence is \QwenTTwoFogEceWorst, \InternTTwoFogEceWorst\ and \GemmaTTwoFogEceWorst\ in fog and \QwenTTwoUwEceWorst, \InternTTwoUwEceWorst\ and \GemmaTTwoUwEceWorst\ under water.
These confidence signals never reach their mid-way level within the sweep, so the blind-spot width is a lower bound (Fig.~\ref{fig:W}, Table~\ref{tab:W}): $W\geq\QwenTTwoFogWInt$, $\InternTTwoFogWInt$ and $\GemmaTTwoFogWInt$ decades in fog for the internal confidence and $\geq\QwenTTwoUwWInt$, $\InternTTwoUwWInt$ and $\GemmaTTwoUwWInt$ under water, i.e. the confidence of the confident models falls at least a factor of \NLagFactorFog\ in visibility later than accuracy in fog and \NLagFactorUw\ under water.
The verbalized part of this overconfidence agrees in direction with earlier reports for proprietary VLMs \cite{groot2024}.
The verbalized confidence has the same lower bounds for these three models; Qwen-4B is the only confident model whose verbalized confidence falls appreciably (\QwenFourTTwoFogVerWorst\ at 1.95\,m in fog, $W=\QwenFourTTwoUwWVer$ [\QwenFourTTwoUwWVerLo, \QwenFourTTwoUwWVerHi] under water), while its internal confidence still does not (\QwenFourTTwoFogIntWorst; $W\geq\QwenFourTTwoFogWInt$).
Smol-2.2B is the exception: its internal confidence falls with accuracy ($W=\SmolTTwoFogWInt$ [\SmolTTwoFogWIntLo, \SmolTTwoFogWIntHi] in fog), but it is low everywhere (\SmolTTwoFogIntMild\ at 62.5\,m, where accuracy is \SmolTTwoFogAccMild), i.e. it is under-confident rather than blind.

\begin{table*}[!t]
\caption{Half-way visibility of accuracy $\Vacc$ (m) and blind-spot width $W$ with 95\,\% intervals for each awareness signal. $\geq$: lower bound (signal never reached its mid-way level). n/a: no parseable verbalized confidence. For Smol-2.2B the main-wording abstain option is chosen on most clean questions, so its $W$ is not interpretable.}
\label{tab:W}
\centering\footnotesize
\resizebox{\textwidth}{!}{\begin{tabular}{@{}lllccccc@{}}
\toprule
Model & Task & Arm & $V_\mathrm{acc}$ (m) & $W$ internal & $W$ verbalized & $W$ abstain & $W$ abstain$_2$ \\
\midrule
Qwen3-VL-2B & T2 & fog & 12.4 & $\geq$0.80 [0.78, 0.82] & $\geq$0.80 [0.78, 0.82] & -0.06 [-0.08, -0.04] & 0.18 [0.15, 0.21] \\
Qwen3-VL-2B & T2 & water & 22.1 & $\geq$1.05 [1.03, 1.08] & $\geq$1.05 [1.03, 1.08] & 0.02 [-0.01, 0.05] & $\geq$1.05 [1.03, 1.08] \\
Qwen3-VL-2B & T3 & fog & 16.8 & $\geq$0.94 [0.91, 0.96] & $\geq$0.94 [0.91, 0.96] & $\geq$0.94 [0.91, 0.96] & $\geq$0.94 [0.91, 0.96] \\
Qwen3-VL-2B & T3 & water & 33.5 & $\geq$1.23 [1.20, 1.27] & $\geq$1.23 [1.20, 1.27] & $\geq$1.23 [1.20, 1.27] & $\geq$1.23 [1.20, 1.27] \\
\midrule
InternVL3.5-2B & T2 & fog & 14.1 & $\geq$0.86 [0.84, 0.88] & $\geq$0.86 [0.84, 0.88] & -0.11 [-0.13, -0.09] & 0.05 [0.03, 0.07] \\
InternVL3.5-2B & T2 & water & 29.5 & $\geq$1.18 [1.16, 1.21] & $\geq$1.18 [1.16, 1.21] & -0.12 [-0.14, -0.09] & 0.10 [0.08, 0.13] \\
InternVL3.5-2B & T3 & fog & 15.9 & $\geq$0.91 [0.89, 0.94] & 0.06 [0.03, 0.09] & $\geq$0.91 [0.89, 0.94] & $\geq$0.91 [0.89, 0.94] \\
InternVL3.5-2B & T3 & water & 34.4 & $\geq$1.25 [1.21, 1.28] & 0.18 [0.14, 0.21] & $\geq$1.25 [1.21, 1.28] & $\geq$1.25 [1.21, 1.28] \\
\midrule
SmolVLM2-2.2B & T2 & fog & 15.9 & 0.07 [0.03, 0.10] & n/a & -0.32 [-0.74, -0.14] & -0.10 [-0.13, -0.06] \\
SmolVLM2-2.2B & T2 & water & 38.6 & 0.06 [0.03, 0.09] & n/a & -0.34 [-0.72, -0.13] & -0.09 [-0.13, -0.06] \\
SmolVLM2-2.2B & T3 & fog & 21.0 & $\geq$1.03 [1.01, 1.06] & n/a & -1.07 [-1.09, -1.02] & $\geq$1.03 [1.01, 1.06] \\
SmolVLM2-2.2B & T3 & water & 58.9 & $\geq$1.48 [1.44, 1.52] & n/a & -0.63 [-0.66, -0.59] & $\geq$1.48 [1.44, 1.52] \\
\midrule
Gemma-4-E2B & T2 & fog & 13.3 & $\geq$0.83 [0.80, 0.87] & $\geq$0.83 [0.80, 0.87] & -0.06 [-0.09, -0.03] & -0.03 [-0.06, -0.00] \\
Gemma-4-E2B & T2 & water & 31.0 & $\geq$1.20 [1.16, 1.25] & $\geq$1.20 [1.16, 1.25] & -0.06 [-0.10, -0.02] & -0.01 [-0.05, 0.03] \\
Gemma-4-E2B & T3 & fog & 19.6 & $\geq$1.00 [0.96, 1.04] & $\geq$1.00 [0.96, 1.04] & $\geq$1.00 [0.96, 1.04] & $\geq$1.00 [0.96, 1.04] \\
Gemma-4-E2B & T3 & water & 49.2 & $\geq$1.40 [1.36, 1.44] & $\geq$1.40 [1.36, 1.44] & $\geq$1.40 [1.36, 1.44] & $\geq$1.40 [1.36, 1.44] \\
\midrule
Qwen3-VL-4B & T2 & fog & 12.1 & $\geq$0.79 [0.76, 0.82] & $\geq$0.79 [0.42, 0.82] & -0.21 [-0.25, -0.18] & -0.06 [-0.09, -0.02] \\
Qwen3-VL-4B & T2 & water & 20.5 & $\geq$1.02 [0.98, 1.06] & 0.75 [0.38, 1.05] & -0.26 [-0.31, -0.22] & -0.05 [-0.09, -0.02] \\
Qwen3-VL-4B & T3 & fog & 15.6 & $\geq$0.90 [0.87, 0.94] & $\geq$0.90 [0.87, 0.94] & $\geq$0.90 [0.87, 0.94] & $\geq$0.90 [0.87, 0.94] \\
Qwen3-VL-4B & T3 & water & 27.2 & $\geq$1.14 [1.11, 1.18] & $\geq$1.14 [1.11, 1.18] & $\geq$1.14 [1.11, 1.18] & $\geq$1.14 [1.11, 1.18] \\
\bottomrule
\end{tabular}
}
\end{table*}

\begin{figure*}[!t]
\centering
\includegraphics[width=\textwidth]{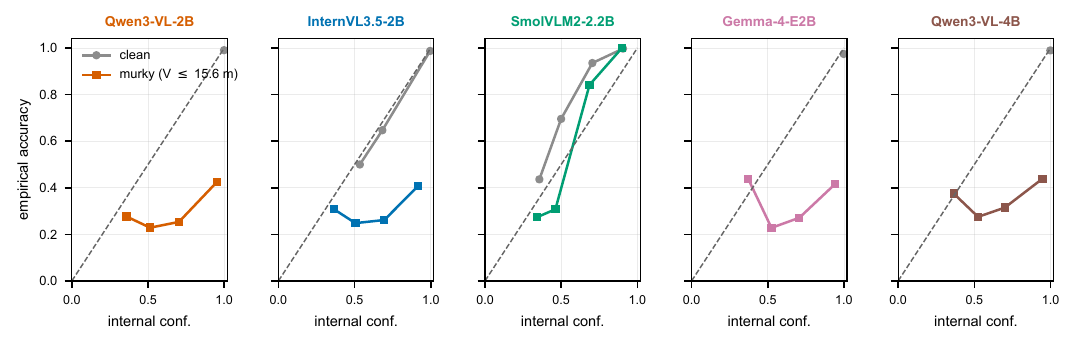}
\caption{Reliability of the internal confidence on \Ttwo, pooled over the murky levels ($V\leq15.6$\,m, fog and underwater) and shown against the clean image.}
\label{fig:rel}
\end{figure*}

The internal confidence discriminates where accuracy is high and loses it in the murky region: the AUROC for detecting errors in fog is \QwenTTwoFogAurocMild, \QwenTTwoFogAurocMid\ and \QwenTTwoFogAurocWorst\ for Qwen-2B at 62.5, 15.6 and 1.95\,m (\InternTTwoFogAurocMild, \InternTTwoFogAurocMid\ and \InternTTwoFogAurocWorst\ for Intern-2B; \GemmaTTwoFogAurocMild, \GemmaTTwoFogAurocMid\ and \GemmaTTwoFogAurocWorst\ for Gemma-E2B).
Pooled over the levels with $V\leq15.6$\,m (both arms), the answers with internal confidence of at least 0.9 are correct only \QwenRelHiAcc\ of the time for Qwen-2B ($n=\QwenRelHiN$), \InternRelHiAcc\ for Intern-2B ($n=\InternRelHiN$) and \GemmaRelHiAcc\ for Gemma-E2B ($n=\GemmaRelHiN$) (Fig.~\ref{fig:rel}); reliability diagrams of this kind are the standard view of calibration \cite{guo2017}.

\subsection{An explicit abstain option helps, but depends on its wording (RQ3)}
\begin{figure*}[!t]
\centering
\includegraphics[width=\textwidth]{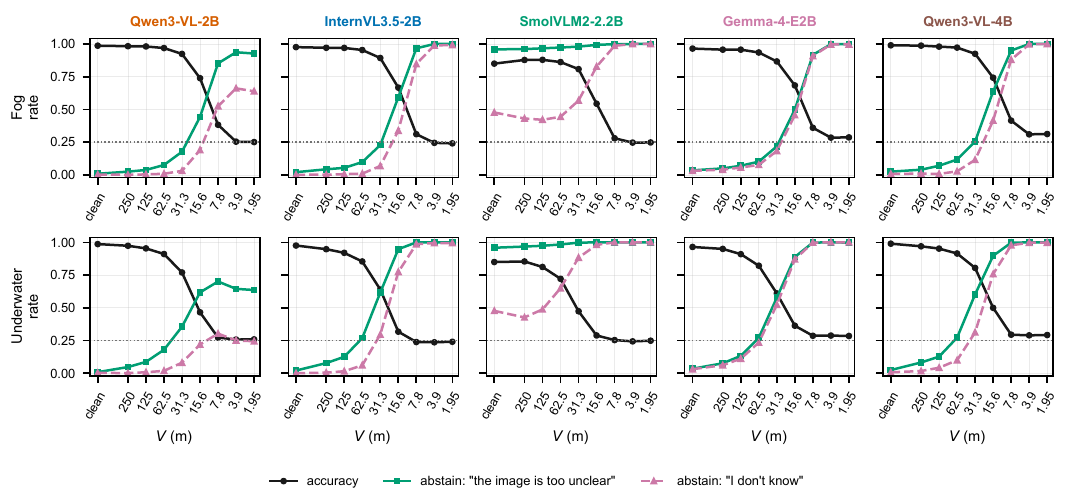}
\caption{Abstain rate on \Ttwo\ under the two wordings of the abstain option, with accuracy for reference.}
\label{fig:wording}
\end{figure*}

With the main wording (``I cannot tell, the image is too unclear to decide'') abstention rises with the loss of accuracy for Qwen-2B, Intern-2B, Gemma-E2B and Qwen-4B (Fig.~\ref{fig:curves}): at 7.8\,m in fog the abstain rate is \QwenTTwoFogAbsDeep, \InternTTwoFogAbsDeep, \GemmaTTwoFogAbsDeep\ and \QwenFourTTwoFogAbsDeep, and its half-way visibility coincides with that of accuracy or is reached earlier ($W$ between \AbsWMinFog\ and \AbsWMaxFog\ in fog and between \AbsWMinUw\ and \AbsWMaxUw\ under water).
With the generic wording ``I don't know'' the result depends on the model (Fig.~\ref{fig:wording}): Qwen-2B abstains on \QwenTTwoFogAbsTwoDeep\ of the questions at 7.8\,m in fog against \QwenTTwoFogAbsDeep\ with the main wording, its abstain$_2$ rate under water never exceeds \QwenTTwoUwAbsTwoDeep\ ($W\geq\QwenTTwoUwWAbsTwo$) and in fog $W=\QwenTTwoFogWAbsTwo$ [\QwenTTwoFogWAbsTwoLo, \QwenTTwoFogWAbsTwoHi]; Intern-2B differs less ($W=\InternTTwoFogWAbsTwo$), and Gemma-E2B and Qwen-4B not at all ($W=\GemmaTTwoFogWAbsTwo$ and $\QwenFourTTwoFogWAbsTwo$ against $\GemmaTTwoFogWAbs$ and $\QwenFourTTwoFogWAbs$ for the main wording).
Smol-2.2B chooses the main-wording option on \SmolTTwoAbsClean\ of the clean questions, so for it only the generic wording is informative (abstain$_2$ rate \SmolTTwoAbsTwoClean\ on clean images, \SmolTTwoFogAbsTwoMild\ at 62.5\,m and \SmolTTwoFogAbsTwoWorst\ at 1.95\,m).
Hence the awareness that an abstain option reveals is conditional on the option naming the \emph{reason} for at least one of the five models (Qwen-2B), and in general on the model.

\subsection{Wording sensitivity of the verbalized confidence}
\label{sec:verbal}
The wording of the verbalized-confidence request was varied on the first 300 images in eight conditions for the two models whose confidence is parseable (Table~\ref{tab:verbalsens}): the main wording (v1), ``rate how likely it is that your answer is correct, from 0 to 100'' (v2) and ``if the image is too unclear to be sure, give a low number'' (v3).
At $V=1.95$\,m in fog the accuracy on \Ttwo\ is \SensQwenTTwoFogTwoAcc\ for Qwen3-VL-2B and \SensInternTTwoFogTwoAcc\ for InternVL3.5-2B, yet the mean verbalized confidence is \SensQwenTTwoFogTwoVOne, \SensQwenTTwoFogTwoVTwo\ and \SensQwenTTwoFogTwoVThree\ for Qwen3-VL-2B and \SensInternTTwoFogTwoVOne, \SensInternTTwoFogTwoVTwo\ and \SensInternTTwoFogTwoVThree\ for InternVL3.5-2B under the three wordings.
The wording moves the level (v2 lowers the confidence of Qwen3-VL-2B to \SensQwenTTwoFogEightVTwo\ at 7.8\,m and raises that of InternVL3.5-2B) but never makes it follow the accuracy, and the instruction that asks for a low number when the image is unclear does not lower it: the verbalized confidence is wording dependent in level but not in tendency.
SmolVLM2 produced no parseable confidence under any wording in the pilot or at stage 1.
\begin{table}[!t]
\caption{Mean verbalized confidence on \Ttwo\ under three wordings of the request (first 300 images; v1 is the main wording restricted to the same images) with the accuracy of the main run for reference.}
\label{tab:verbalsens}
\centering\scriptsize
\resizebox{\columnwidth}{!}{\begin{tabular}{@{}llcccc@{}}
\toprule
Model & Condition & Accuracy & v1 (main) & v2 & v3 \\
\midrule
Qwen3-VL-2B & clean & 0.99 & 0.95 & 0.99 & 0.98 \\
Qwen3-VL-2B & fog 62.5 & 0.98 & 0.95 & 0.98 & 0.96 \\
Qwen3-VL-2B & fog 15.6 & 0.74 & 0.93 & 0.84 & 0.93 \\
Qwen3-VL-2B & fog 7.8 & 0.36 & 0.89 & 0.76 & 0.92 \\
Qwen3-VL-2B & fog 1.95 & 0.22 & 0.90 & 0.73 & 0.96 \\
Qwen3-VL-2B & water 31.3 & 0.77 & 0.92 & 0.90 & 0.93 \\
Qwen3-VL-2B & water 7.8 & 0.23 & 0.90 & 0.83 & 0.95 \\
Qwen3-VL-2B & water 1.95 & 0.21 & 0.90 & 0.84 & 0.95 \\
\midrule
InternVL3.5-2B & clean & 0.96 & 0.90 & 0.98 & 0.90 \\
InternVL3.5-2B & fog 62.5 & 0.94 & 0.90 & 0.97 & 0.90 \\
InternVL3.5-2B & fog 15.6 & 0.69 & 0.90 & 0.95 & 0.90 \\
InternVL3.5-2B & fog 7.8 & 0.30 & 0.92 & 0.97 & 0.90 \\
InternVL3.5-2B & fog 1.95 & 0.26 & 0.94 & 0.99 & 0.90 \\
InternVL3.5-2B & water 31.3 & 0.64 & 0.90 & 0.94 & 0.90 \\
InternVL3.5-2B & water 7.8 & 0.25 & 0.92 & 0.97 & 0.91 \\
InternVL3.5-2B & water 1.95 & 0.26 & 0.93 & 0.98 & 0.91 \\
\bottomrule
\end{tabular}
}
\end{table}

\subsection{Nominal severities do not reach the failure band (RQ4)}
\begin{table*}[!t]
\caption{ImageNet-C fog severity expressed as physical visibility (no VLM involved) and \Ttwo\ accuracy at that severity. $V_\mathrm{eq}$: fog visibility with the same retained contrast gain, median and interquartile range over 1000 images; P90/P10: spread across images.}
\label{tab:nominal}
\centering\footnotesize
\setlength{\tabcolsep}{4pt}
\resizebox{0.9\textwidth}{!}{\begin{tabular}{@{}cccccccccc@{}}
\toprule
Severity & Mean gain & Median $V_\mathrm{eq}$ (m) & IQR (m) & P90/P10 & Qwen3-VL-2B & InternVL3.5-2B & SmolVLM2-2.2B & Gemma-4-E2B & Qwen3-VL-4B \\
\midrule
1 & 0.40 & 110 & 70--174 & 6.2$\times$ & 0.98 & 0.97 & 0.89 & 0.96 & 0.98 \\
2 & 0.33 & 82 & 51--132 & 6.9$\times$ & 0.98 & 0.96 & 0.89 & 0.96 & 0.99 \\
3 & 0.28 & 67 & 40--112 & 7.8$\times$ & 0.98 & 0.96 & 0.89 & 0.96 & 0.98 \\
4 & 0.28 & 65 & 40--112 & 7.7$\times$ & 0.98 & 0.96 & 0.87 & 0.96 & 0.97 \\
5 & 0.24 & 55 & 33--96 & 8.4$\times$ & 0.97 & 0.96 & 0.85 & 0.94 & 0.97 \\
\bottomrule
\end{tabular}
}
\end{table*}
Table~\ref{tab:nominal} translates ImageNet-C fog severities 1--5 into visibility.
By retained contrast gain they correspond to a median $V_\mathrm{eq}$ of \NomVeqOne, \NomVeqTwo, \NomVeqThree, \NomVeqFour\ and \NomVeqFive\,m, with a P90/P10 spread across images of \NomSpreadOne--\NomSpreadFive$\times$; severities 3 and 4 share the same amplitude constant in the implementation and differ only in the roughness of the fractal.
The statistic frozen before the pilot, the ratio of luma standard deviations, gave \NomStdVeqOne, \NomStdVeqTwo, \NomStdVeqThree, \NomStdVeqFour\ and \NomStdVeqFive\,m, which does not decrease with severity because the additive haze adds variance.
Only \NomMonoFrac\,\% of the images have a gain-matched $V_\mathrm{eq}$ that decreases monotonically over severities 1--5.

\begin{figure*}[!t]
\centering
\includegraphics[width=\textwidth]{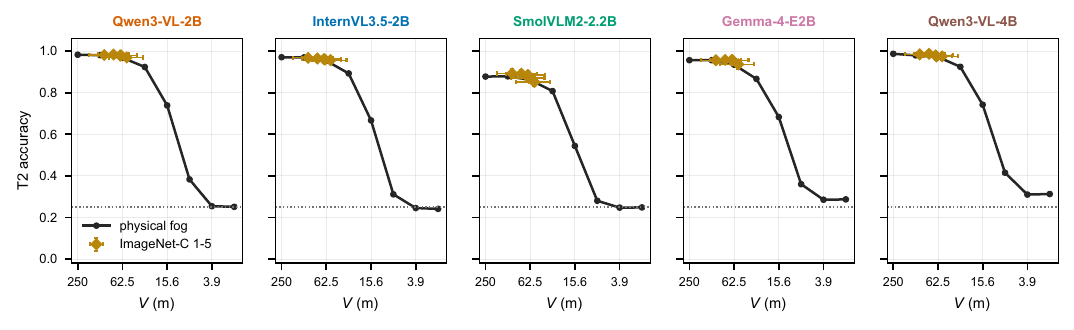}
\caption{Physical fog curve (\Ttwo\ accuracy) with the five ImageNet-C severities placed at their median equivalent visibility (horizontal bars: interquartile range of $V_\mathrm{eq}$ over images; vertical bars: 95\,\% CI of accuracy).}
\label{fig:nominal}
\end{figure*}
Nominal severities therefore sit on the plateau (Fig.~\ref{fig:nominal}): \Ttwo\ accuracy over severities 1--5 spans \QwenTTwoNomMin--\QwenTTwoNomMax\ for Qwen-2B, \InternTTwoNomMin--\InternTTwoNomMax\ for Intern-2B, \GemmaTTwoNomMin--\GemmaTTwoNomMax\ for Gemma-E2B, \QwenFourTTwoNomMin--\QwenFourTTwoNomMax\ for Qwen-4B and \SmolTTwoNomMin--\SmolTTwoNomMax\ for Smol-2.2B (clean \SmolTTwoAccClean), while the physical sweep covers the whole range down to chance.
The nominal series are monotone within noise but uninformative: pooled over models and tasks, the dynamic range of accuracy is \MonoNomAccRng\ percentage points against \MonoFogAccRng\ for the physical fog sweep.

\subsection{Verification task and controls}
\label{sec:diag}
\begin{figure*}[!t]
\centering
\includegraphics[width=\textwidth]{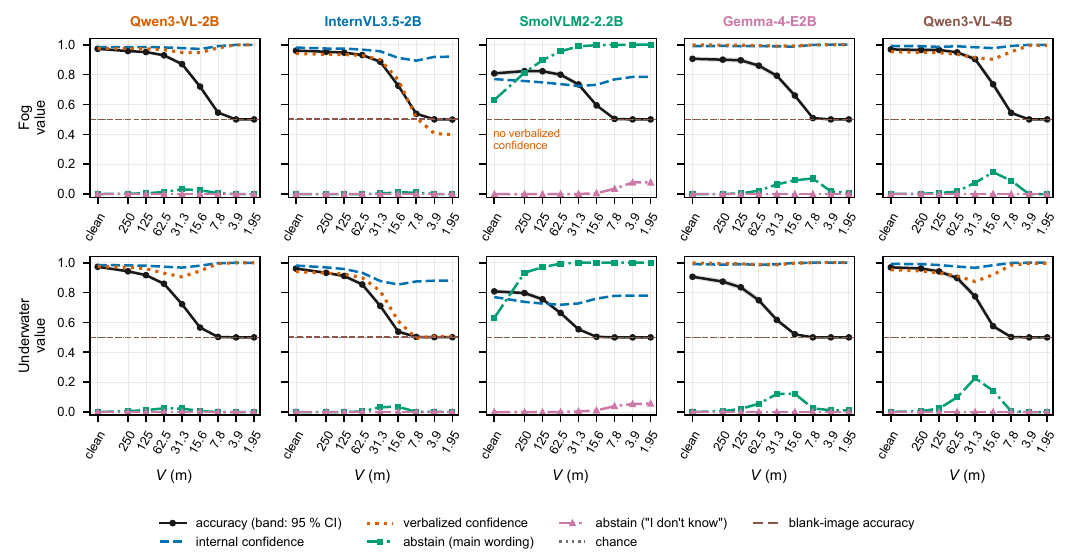}
\caption{Verification task (\Tthree): accuracy and read-outs against $V$; same conventions as Fig.~\ref{fig:curves}. Chance is 0.5.}
\label{fig:curvesT3}
\end{figure*}

Figure~\ref{fig:curvesT3} shows the verification task \Tthree\ with the same conventions as Fig.~\ref{fig:curves}; its blind-spot widths are in the lower half of Fig.~\ref{fig:W} and in Table~\ref{tab:W}.
\begin{table*}[!t]
\caption{Image controls. Blank: uniform grey image (box kept). Shuffled: clean pixels of another image. Int.\ conf., verb.\ conf.\ and abstain rates are for the blank image.}
\label{tab:controls}
\centering\footnotesize
\resizebox{0.9\textwidth}{!}{\begin{tabular}{@{}llccccccc@{}}
\toprule
Model & Task & Clean & Blank & Shuffled & Blank int.\ conf. & Blank verb.\ conf. & Blank abst. & Blank abst.$_2$ \\
\midrule
Qwen3-VL-2B & T2 & 0.99 & 0.26 & 0.29 & 0.82 & 0.90 & 0.85 & 0.45 \\
Qwen3-VL-2B & T3 & 0.97 & 0.50 & 0.51 & 1.00 & 1.00 & 0.00 & 0.00 \\
InternVL3.5-2B & T2 & 0.98 & 0.24 & 0.29 & 0.69 & 0.92 & 1.00 & 0.99 \\
InternVL3.5-2B & T3 & 0.96 & 0.50 & 0.52 & 0.92 & 0.34 & 0.00 & 0.00 \\
SmolVLM2-2.2B & T2 & 0.85 & 0.25 & 0.25 & 0.38 & n/a & 1.00 & 1.00 \\
SmolVLM2-2.2B & T3 & 0.81 & 0.50 & 0.50 & 0.78 & n/a & 1.00 & 0.05 \\
Gemma-4-E2B & T2 & 0.96 & 0.28 & 0.27 & 0.80 & 0.99 & 1.00 & 1.00 \\
Gemma-4-E2B & T3 & 0.91 & 0.50 & 0.51 & 1.00 & 1.00 & 0.03 & 0.00 \\
Qwen3-VL-4B & T2 & 0.99 & 0.30 & 0.29 & 0.84 & 0.65 & 1.00 & 0.99 \\
Qwen3-VL-4B & T3 & 0.97 & 0.50 & 0.51 & 1.00 & 0.99 & 0.00 & 0.00 \\
\bottomrule
\end{tabular}
}
\end{table*}
\emph{(a) The floor of \Ttwo\ is chance.}
Accuracy on the blank image is \QwenTTwoBlank, \InternTTwoBlank, \GemmaTTwoBlank, \QwenFourTTwoBlank\ and \SmolTTwoBlank\ for Qwen-2B, Intern-2B, Gemma-E2B, Qwen-4B and Smol-2.2B, and on a shuffled image \QwenTTwoShuf, \InternTTwoShuf, \GemmaTTwoShuf, \QwenFourTTwoShuf\ and \SmolTTwoShuf\ (chance 0.25), so after prior matching the identification task cannot be solved from the box and the category prior alone; the pilot's below-chance floor (0.11--0.16) is gone.
\emph{(b) \Tthree\ is dominated by an answer bias.}
On the blank image the models answer ``No'' to (almost) every question (accuracy \QwenTThreeBlank, \InternTThreeBlank, \GemmaTThreeBlank\ and \SmolTThreeBlank), and at 1.95\,m in fog the accuracy on the true-category questions is \QwenTThreeYesFogWorst, \InternTThreeYesFogWorst\ and \GemmaTThreeYesFogWorst\ against \QwenTThreeNoFogWorst, \InternTThreeNoFogWorst\ and \GemmaTThreeNoFogWorst\ on the false-category questions.
The accuracy of \Tthree\ therefore falls to 0.5 because ``No'' becomes the only answer; its internal confidence stays at \QwenTThreeFogIntWorst, \InternTThreeFogIntWorst\ and \GemmaTThreeFogIntWorst; it measures whether the boxed object is judged visible rather than how well the answer is calibrated.
\emph{(c) The abstain option is almost unused on \Tthree\ by the confident models (at most \QwenTThreeFogAbsDeep, \InternTThreeFogAbsDeep, \GemmaTThreeFogAbsDeep\ and \QwenFourTThreeFogAbsDeep\ of the questions at 7.8\,m)}, in contrast to \Ttwo.

\subsection{Complete sweep}
\label{sec:levels}
Tables~\ref{tab:levelsT2} and~\ref{tab:levelsT3} list accuracy, internal and verbalized confidence and the abstain rates at every level of the sweep for each model and arm; they are the data behind Figs.~\ref{fig:curves} and~\ref{fig:curvesT3}.
\begin{table*}[!t]
\caption{Identification task (\Ttwo): all signals at every visibility level ($V$ in metres). ``--'': not available (verbalized confidence of SmolVLM2).}
\label{tab:levelsT2}
\centering\scriptsize
\resizebox{\textwidth}{!}{\begin{tabular}{@{}lllccccccccc@{}}
\toprule
Model & Arm & Signal & clean & 250 & 125 & 62.5 & 31.25 & 15.625 & 7.8125 & 3.90625 & 1.95312 \\
\midrule
Qwen3-VL-2B & fog & accuracy & 0.99 & 0.98 & 0.98 & 0.97 & 0.92 & 0.74 & 0.38 & 0.25 & 0.25 \\
 &  & internal conf. & 1.00 & 0.99 & 0.99 & 0.99 & 0.97 & 0.91 & 0.82 & 0.80 & 0.81 \\
 &  & verbalized conf. & 0.95 & 0.95 & 0.95 & 0.95 & 0.94 & 0.93 & 0.90 & 0.90 & 0.90 \\
 &  & abstain & 0.01 & 0.03 & 0.04 & 0.08 & 0.18 & 0.44 & 0.85 & 0.94 & 0.93 \\
 &  & abstain$_2$ & 0.00 & 0.00 & 0.00 & 0.01 & 0.03 & 0.19 & 0.53 & 0.66 & 0.64 \\
Qwen3-VL-2B & water & accuracy & 0.99 & 0.97 & 0.95 & 0.91 & 0.77 & 0.47 & 0.27 & 0.26 & 0.26 \\
 &  & internal conf. & 1.00 & 0.99 & 0.99 & 0.97 & 0.92 & 0.83 & 0.81 & 0.82 & 0.82 \\
 &  & verbalized conf. & 0.95 & 0.95 & 0.95 & 0.94 & 0.93 & 0.90 & 0.90 & 0.90 & 0.90 \\
 &  & abstain & 0.01 & 0.05 & 0.09 & 0.18 & 0.36 & 0.62 & 0.70 & 0.65 & 0.64 \\
 &  & abstain$_2$ & 0.00 & 0.00 & 0.01 & 0.02 & 0.08 & 0.22 & 0.30 & 0.25 & 0.24 \\
\midrule
InternVL3.5-2B & fog & accuracy & 0.98 & 0.97 & 0.97 & 0.95 & 0.89 & 0.67 & 0.31 & 0.24 & 0.24 \\
 &  & internal conf. & 0.98 & 0.97 & 0.97 & 0.97 & 0.93 & 0.82 & 0.68 & 0.68 & 0.68 \\
 &  & verbalized conf. & 0.90 & 0.90 & 0.90 & 0.90 & 0.90 & 0.90 & 0.92 & 0.93 & 0.93 \\
 &  & abstain & 0.02 & 0.04 & 0.05 & 0.10 & 0.23 & 0.59 & 0.97 & 1.00 & 1.00 \\
 &  & abstain$_2$ & 0.00 & 0.00 & 0.01 & 0.01 & 0.07 & 0.34 & 0.85 & 0.99 & 0.99 \\
InternVL3.5-2B & water & accuracy & 0.98 & 0.95 & 0.92 & 0.85 & 0.64 & 0.32 & 0.24 & 0.24 & 0.24 \\
 &  & internal conf. & 0.98 & 0.96 & 0.95 & 0.90 & 0.77 & 0.67 & 0.67 & 0.68 & 0.68 \\
 &  & verbalized conf. & 0.90 & 0.90 & 0.90 & 0.90 & 0.90 & 0.91 & 0.92 & 0.93 & 0.93 \\
 &  & abstain & 0.02 & 0.08 & 0.13 & 0.27 & 0.62 & 0.95 & 1.00 & 1.00 & 1.00 \\
 &  & abstain$_2$ & 0.00 & 0.01 & 0.02 & 0.06 & 0.30 & 0.77 & 0.99 & 0.99 & 0.99 \\
\midrule
SmolVLM2-2.2B & fog & accuracy & 0.85 & 0.88 & 0.88 & 0.86 & 0.81 & 0.54 & 0.28 & 0.25 & 0.25 \\
 &  & internal conf. & 0.70 & 0.72 & 0.72 & 0.70 & 0.63 & 0.49 & 0.39 & 0.38 & 0.38 \\
 &  & verbalized conf. & -- & -- & -- & -- & -- & -- & -- & -- & -- \\
 &  & abstain & 0.96 & 0.96 & 0.97 & 0.97 & 0.98 & 0.99 & 1.00 & 1.00 & 1.00 \\
 &  & abstain$_2$ & 0.48 & 0.43 & 0.42 & 0.44 & 0.57 & 0.83 & 0.99 & 1.00 & 1.00 \\
SmolVLM2-2.2B & water & accuracy & 0.85 & 0.85 & 0.81 & 0.72 & 0.47 & 0.29 & 0.25 & 0.24 & 0.25 \\
 &  & internal conf. & 0.70 & 0.70 & 0.66 & 0.57 & 0.46 & 0.39 & 0.38 & 0.38 & 0.38 \\
 &  & verbalized conf. & -- & -- & -- & -- & -- & -- & -- & -- & -- \\
 &  & abstain & 0.96 & 0.97 & 0.97 & 0.98 & 1.00 & 1.00 & 1.00 & 1.00 & 1.00 \\
 &  & abstain$_2$ & 0.48 & 0.43 & 0.49 & 0.65 & 0.88 & 0.98 & 1.00 & 1.00 & 1.00 \\
\midrule
Gemma-4-E2B & fog & accuracy & 0.96 & 0.96 & 0.96 & 0.94 & 0.87 & 0.68 & 0.36 & 0.28 & 0.29 \\
 &  & internal conf. & 1.00 & 1.00 & 0.99 & 0.99 & 0.98 & 0.92 & 0.82 & 0.79 & 0.79 \\
 &  & verbalized conf. & 1.00 & 1.00 & 1.00 & 1.00 & 0.99 & 0.99 & 0.99 & 1.00 & 1.00 \\
 &  & abstain & 0.04 & 0.05 & 0.07 & 0.10 & 0.22 & 0.50 & 0.92 & 1.00 & 1.00 \\
 &  & abstain$_2$ & 0.03 & 0.04 & 0.06 & 0.08 & 0.18 & 0.46 & 0.91 & 1.00 & 1.00 \\
Gemma-4-E2B & water & accuracy & 0.96 & 0.95 & 0.91 & 0.82 & 0.61 & 0.36 & 0.29 & 0.29 & 0.28 \\
 &  & internal conf. & 1.00 & 0.99 & 0.98 & 0.97 & 0.91 & 0.84 & 0.80 & 0.80 & 0.80 \\
 &  & verbalized conf. & 1.00 & 1.00 & 1.00 & 0.99 & 0.98 & 0.98 & 0.99 & 0.99 & 0.99 \\
 &  & abstain & 0.04 & 0.08 & 0.13 & 0.27 & 0.58 & 0.89 & 1.00 & 1.00 & 1.00 \\
 &  & abstain$_2$ & 0.03 & 0.06 & 0.11 & 0.23 & 0.52 & 0.87 & 1.00 & 1.00 & 1.00 \\
\midrule
Qwen3-VL-4B & fog & accuracy & 0.99 & 0.99 & 0.98 & 0.97 & 0.93 & 0.74 & 0.41 & 0.31 & 0.31 \\
 &  & internal conf. & 1.00 & 1.00 & 1.00 & 0.99 & 0.98 & 0.92 & 0.85 & 0.84 & 0.84 \\
 &  & verbalized conf. & 0.95 & 0.95 & 0.95 & 0.94 & 0.92 & 0.85 & 0.72 & 0.62 & 0.61 \\
 &  & abstain & 0.03 & 0.04 & 0.07 & 0.12 & 0.26 & 0.64 & 0.95 & 1.00 & 1.00 \\
 &  & abstain$_2$ & 0.01 & 0.01 & 0.01 & 0.03 & 0.12 & 0.41 & 0.88 & 1.00 & 1.00 \\
Qwen3-VL-4B & water & accuracy & 0.99 & 0.97 & 0.95 & 0.92 & 0.81 & 0.50 & 0.29 & 0.29 & 0.29 \\
 &  & internal conf. & 1.00 & 0.99 & 0.99 & 0.97 & 0.93 & 0.86 & 0.83 & 0.82 & 0.82 \\
 &  & verbalized conf. & 0.95 & 0.94 & 0.94 & 0.92 & 0.86 & 0.75 & 0.63 & 0.60 & 0.59 \\
 &  & abstain & 0.03 & 0.08 & 0.13 & 0.27 & 0.60 & 0.90 & 1.00 & 1.00 & 1.00 \\
 &  & abstain$_2$ & 0.01 & 0.02 & 0.04 & 0.10 & 0.31 & 0.76 & 0.98 & 1.00 & 1.00 \\
\bottomrule
\end{tabular}
}
\end{table*}

\begin{table*}[!t]
\caption{Verification task (\Tthree): all signals at every visibility level.}
\label{tab:levelsT3}
\centering\scriptsize
\resizebox{\textwidth}{!}{\begin{tabular}{@{}lllccccccccc@{}}
\toprule
Model & Arm & Signal & clean & 250 & 125 & 62.5 & 31.25 & 15.625 & 7.8125 & 3.90625 & 1.95312 \\
\midrule
Qwen3-VL-2B & fog & accuracy & 0.97 & 0.96 & 0.95 & 0.93 & 0.87 & 0.72 & 0.55 & 0.50 & 0.50 \\
 &  & internal conf. & 0.98 & 0.98 & 0.98 & 0.98 & 0.98 & 0.97 & 0.99 & 1.00 & 1.00 \\
 &  & verbalized conf. & 0.97 & 0.97 & 0.97 & 0.97 & 0.95 & 0.95 & 0.98 & 1.00 & 1.00 \\
 &  & abstain & 0.00 & 0.00 & 0.01 & 0.01 & 0.03 & 0.03 & 0.01 & 0.00 & 0.00 \\
 &  & abstain$_2$ & 0.00 & 0.00 & 0.00 & 0.00 & 0.00 & 0.00 & 0.00 & 0.00 & 0.00 \\
Qwen3-VL-2B & water & accuracy & 0.97 & 0.94 & 0.92 & 0.86 & 0.72 & 0.57 & 0.50 & 0.50 & 0.50 \\
 &  & internal conf. & 0.98 & 0.98 & 0.98 & 0.97 & 0.97 & 0.98 & 1.00 & 1.00 & 1.00 \\
 &  & verbalized conf. & 0.97 & 0.97 & 0.96 & 0.93 & 0.91 & 0.95 & 0.99 & 1.00 & 1.00 \\
 &  & abstain & 0.00 & 0.01 & 0.01 & 0.03 & 0.02 & 0.01 & 0.00 & 0.00 & 0.00 \\
 &  & abstain$_2$ & 0.00 & 0.00 & 0.00 & 0.00 & 0.00 & 0.00 & 0.00 & 0.00 & 0.00 \\
\midrule
InternVL3.5-2B & fog & accuracy & 0.96 & 0.95 & 0.95 & 0.93 & 0.89 & 0.73 & 0.54 & 0.50 & 0.50 \\
 &  & internal conf. & 0.98 & 0.97 & 0.97 & 0.97 & 0.96 & 0.91 & 0.89 & 0.92 & 0.92 \\
 &  & verbalized conf. & 0.94 & 0.94 & 0.93 & 0.93 & 0.90 & 0.76 & 0.51 & 0.41 & 0.40 \\
 &  & abstain & 0.00 & 0.00 & 0.00 & 0.00 & 0.00 & 0.01 & 0.01 & 0.00 & 0.00 \\
 &  & abstain$_2$ & 0.00 & 0.00 & 0.00 & 0.00 & 0.00 & 0.00 & 0.00 & 0.00 & 0.00 \\
InternVL3.5-2B & water & accuracy & 0.96 & 0.93 & 0.91 & 0.85 & 0.71 & 0.54 & 0.50 & 0.50 & 0.50 \\
 &  & internal conf. & 0.98 & 0.97 & 0.96 & 0.93 & 0.88 & 0.85 & 0.87 & 0.88 & 0.88 \\
 &  & verbalized conf. & 0.94 & 0.93 & 0.92 & 0.90 & 0.81 & 0.61 & 0.49 & 0.51 & 0.51 \\
 &  & abstain & 0.00 & 0.00 & 0.00 & 0.00 & 0.03 & 0.04 & 0.00 & 0.00 & 0.00 \\
 &  & abstain$_2$ & 0.00 & 0.00 & 0.00 & 0.00 & 0.00 & 0.00 & 0.00 & 0.00 & 0.00 \\
\midrule
SmolVLM2-2.2B & fog & accuracy & 0.81 & 0.82 & 0.82 & 0.80 & 0.73 & 0.59 & 0.50 & 0.50 & 0.50 \\
 &  & internal conf. & 0.77 & 0.76 & 0.75 & 0.74 & 0.72 & 0.73 & 0.77 & 0.78 & 0.79 \\
 &  & verbalized conf. & -- & -- & -- & -- & -- & -- & -- & -- & -- \\
 &  & abstain & 0.63 & 0.81 & 0.90 & 0.96 & 0.99 & 1.00 & 1.00 & 1.00 & 1.00 \\
 &  & abstain$_2$ & 0.00 & 0.00 & 0.00 & 0.00 & 0.00 & 0.01 & 0.04 & 0.08 & 0.08 \\
SmolVLM2-2.2B & water & accuracy & 0.81 & 0.80 & 0.76 & 0.66 & 0.55 & 0.50 & 0.50 & 0.50 & 0.50 \\
 &  & internal conf. & 0.77 & 0.74 & 0.73 & 0.72 & 0.73 & 0.76 & 0.78 & 0.78 & 0.78 \\
 &  & verbalized conf. & -- & -- & -- & -- & -- & -- & -- & -- & -- \\
 &  & abstain & 0.63 & 0.93 & 0.97 & 0.99 & 1.00 & 1.00 & 1.00 & 1.00 & 1.00 \\
 &  & abstain$_2$ & 0.00 & 0.00 & 0.00 & 0.00 & 0.00 & 0.01 & 0.04 & 0.05 & 0.06 \\
\midrule
Gemma-4-E2B & fog & accuracy & 0.91 & 0.90 & 0.90 & 0.86 & 0.79 & 0.66 & 0.51 & 0.50 & 0.50 \\
 &  & internal conf. & 0.99 & 0.99 & 0.99 & 0.99 & 0.99 & 0.99 & 1.00 & 1.00 & 1.00 \\
 &  & verbalized conf. & 1.00 & 1.00 & 1.00 & 0.99 & 0.99 & 0.99 & 1.00 & 1.00 & 1.00 \\
 &  & abstain & 0.00 & 0.00 & 0.01 & 0.02 & 0.07 & 0.09 & 0.10 & 0.02 & 0.01 \\
 &  & abstain$_2$ & 0.00 & 0.00 & 0.00 & 0.00 & 0.00 & 0.00 & 0.00 & 0.00 & 0.00 \\
Gemma-4-E2B & water & accuracy & 0.91 & 0.87 & 0.83 & 0.75 & 0.62 & 0.52 & 0.50 & 0.50 & 0.50 \\
 &  & internal conf. & 0.99 & 0.99 & 0.99 & 0.99 & 0.99 & 1.00 & 1.00 & 1.00 & 1.00 \\
 &  & verbalized conf. & 1.00 & 0.99 & 0.99 & 0.99 & 0.99 & 1.00 & 1.00 & 1.00 & 1.00 \\
 &  & abstain & 0.00 & 0.01 & 0.02 & 0.06 & 0.12 & 0.12 & 0.03 & 0.01 & 0.01 \\
 &  & abstain$_2$ & 0.00 & 0.00 & 0.00 & 0.00 & 0.00 & 0.00 & 0.00 & 0.00 & 0.00 \\
\midrule
Qwen3-VL-4B & fog & accuracy & 0.97 & 0.96 & 0.96 & 0.95 & 0.90 & 0.73 & 0.54 & 0.50 & 0.50 \\
 &  & internal conf. & 0.99 & 0.99 & 0.99 & 0.99 & 0.98 & 0.98 & 0.99 & 1.00 & 1.00 \\
 &  & verbalized conf. & 0.95 & 0.95 & 0.95 & 0.94 & 0.91 & 0.90 & 0.95 & 0.99 & 1.00 \\
 &  & abstain & 0.00 & 0.00 & 0.01 & 0.02 & 0.08 & 0.15 & 0.09 & 0.00 & 0.00 \\
 &  & abstain$_2$ & 0.00 & 0.00 & 0.00 & 0.00 & 0.00 & 0.00 & 0.00 & 0.00 & 0.00 \\
Qwen3-VL-4B & water & accuracy & 0.97 & 0.96 & 0.94 & 0.90 & 0.77 & 0.57 & 0.50 & 0.50 & 0.50 \\
 &  & internal conf. & 0.99 & 0.99 & 0.98 & 0.97 & 0.97 & 0.98 & 1.00 & 1.00 & 1.00 \\
 &  & verbalized conf. & 0.95 & 0.94 & 0.93 & 0.91 & 0.87 & 0.92 & 0.98 & 1.00 & 1.00 \\
 &  & abstain & 0.00 & 0.01 & 0.02 & 0.10 & 0.23 & 0.14 & 0.01 & 0.00 & 0.00 \\
 &  & abstain$_2$ & 0.00 & 0.00 & 0.00 & 0.00 & 0.00 & 0.00 & 0.00 & 0.00 & 0.00 \\
\bottomrule
\end{tabular}
}
\end{table*}

\subsection{Qualitative examples}
\begin{figure*}[!t]
\centering
\includegraphics[width=0.96\textwidth]{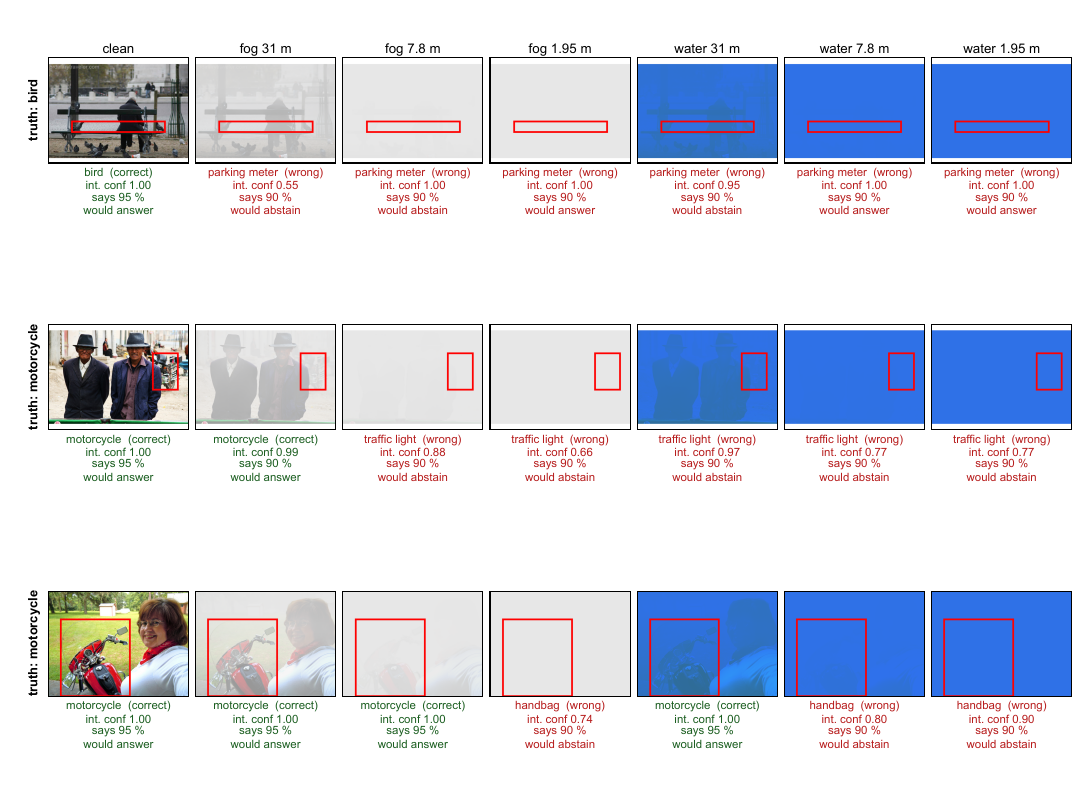}
\caption{Qualitative results of Qwen3-VL-2B on \Ttwo\ for the first three images of the subset. Green: correct; red: wrong. ``int.\ conf.'': internal confidence; ``says'': verbalized confidence; the last line is the decision when the abstain option is offered (main wording).}
\label{fig:qual}
\end{figure*}

Figure~\ref{fig:qual} shows the first three images of the subset (not selected) across fog and underwater levels with Qwen-2B's answer, internal and verbalized confidence, and the abstain decision.

\section{Discussion}
\label{sec:discussion}

\paragraph*{Knowing versus saying}
The read-outs disagree in a systematic way.
The internal confidence still carries information where accuracy is high (AUROC close to one at mild fog), so the models are not blind in a representational sense; but its mean level barely moves while accuracy collapses, and at the murkiest levels its discrimination is gone.
The verbalized confidence is nearly constant on the identification task for three of the four models that produce it, in agreement with the report that the stated confidence of a small VLM stays almost flat under degradation \cite{ferdous2026}; the exception is the largest model, Qwen3-VL-4B, whose verbalized confidence falls to \QwenFourTTwoFogVerWorst\ although its internal confidence does not (\QwenFourTTwoFogIntWorst), so with scale the stated confidence may become the more informative signal, which a larger study should test.
This differs from the Qwen2-VL family, where scale improved the internal signal and left the verbalized one weak \cite{ferdous2026b}; the studies differ in model family, degradations and metric (a mean level here, an AUROC there), and calibration is known to depend on architecture beyond size \cite{minderer2021}, so we do not read the difference as a scale effect.
For language models tuned with human feedback, verbalized confidence has been found better calibrated than token probabilities \cite{tian2023}, and a model can be taught to verbalize calibrated uncertainty \cite{lin2022words}; our tasks, models and elicitation differ, so we do not claim a contradiction, only that on this axis and for these small VLMs it is not a reliable monitor.

\paragraph*{An abstain option is a cue, not a capability}
With the main wording the abstain rate follows the loss of accuracy ($W$ between \AbsWMinFog\ and \AbsWMaxFog\ in fog), and with ``I don't know'' it does not ($W$ up to \AbsTwoWMaxFog\ in fog; under water Qwen3-VL-2B's generic abstain rate stays at \QwenTTwoUwAbsTwoWorst).
The awareness we measure with an abstain option is therefore conditional on the option naming the right reason.
Abstention in language models is usually trained for \cite{zhang2024rtuning} and depends on the query and the model as well as on human values \cite{wen2025abstain}; in deployment a cue of the kind we tested would have to be part of the prompt, and its wording becomes a calibration parameter.
This is consistent with the finding that VLMs rarely abstain under standard prompting \cite{madhusudhan2026}.

\paragraph*{What a physical axis adds}
On a nominal scale the ImageNet-C fog severities 1--5 leave these models at nearly their clean accuracy (within \NomMaxDevPP\ percentage points), and a benchmark built on them would conclude that they are robust to fog.
On the physical axis the same models are unaffected down to about 60\,m and then fail within a factor of ten in visibility.
The two statements are compatible because the nominal severities are equivalent to \NomVeqFive--\NomVeqOne\,m and differ from image to image by a factor of \NomSpreadOne--\NomSpreadFive.
For a user, ``fails below about \VaccFogMin--\VaccFogMax\,m of visibility, under the assumed depth scale'' is a more actionable statement than ``robust at severity 5''.
A visibility range can also be estimated from a camera image \cite{hautiere2006}, which connects such a statement to a quantity that a deployed system can measure.
The scale-free $W$ additionally allows comparing models without trusting the depth estimate.

\paragraph*{Fog versus underwater}
Under water the failure begins at a larger best-channel visibility ($\Vacc$ of \VaccUwMin--\VaccUwMax\,m) and accuracy again ends at chance.
The lower bounds on $W$ should \emph{not} be compared across arms: with the sweep ending at 1.95\,m, a larger $\Vacc$ mechanically gives a larger bound.
Within the limits of the clear-water arm, the confidence gap at the end of the sweep is similar in the two arms (ECE of the internal confidence \QwenTTwoUwEceWorst\ and \InternTTwoUwEceWorst\ under water against \QwenTTwoFogEceWorst\ and \InternTTwoFogEceWorst\ in fog).

\paragraph*{Practical reading}
For a system that relies on a small VLM in fog or turbid water, these results suggest not trusting (i) the verbalized confidence of the 2--2.2\,B models, (ii) internal confidence without a calibration at the relevant visibility, and (iii) a generic ``I don't know'' option.
Real users do supply images for which a question cannot be answered \cite{gurari2018vizwiz}, and in a degraded or missing image a VLM may still produce a plausible answer from general knowledge \cite{xie2025drivebench}.
A cheaper and more robust monitor would be a separate image-quality or visibility estimate that triggers abstention before the VLM is asked; testing that is outside this study.

\section{Limitations}
\label{sec:limits}

\begin{itemize}
\item \emph{Scale.} 1000 images, one image sample (the same images for all models, drawn once), five open models of 2--5\,B parameters (Gemma-4-E2B on the first \GemmaNImg\ images and Qwen3-VL-4B on the first \QwenFourNImg\ only, for time), one prompt family. The intervals reflect image sampling only. Other families, larger models and repeated image samples may behave differently.
\item \emph{Depth.} Monocular metric depth of unverified scale; visibilities in metres are conditional on it (Section~\ref{sec:method}); $W$ is not. Depth errors that are not a global scale (sky, reflective surfaces) are not modelled, and depth estimators are themselves affected by weather corruptions \cite{kong2023robodepth}.
\item \emph{Synthetic physics.} Eq.~(\ref{eq:formation}) is applied to photographs of in-air scenes with one homogeneous medium. Real fog has been captured in driving datasets with correspondences between adverse and clear conditions \cite{sakaridis2021acdc}, which this study does not use. The underwater arm uses only the clear-ocean spectral shape (absorption of pure water, no particles, no scattering, monochromatic channels); water types differ in their spectral profiles \cite{berman2021}, so it is not representative of coastal or turbid water. Forward scattering and blur are not modelled.
\item \emph{Axes are not comparable across arms} in absolute metres (different defining channels), and lower bounds on $W$ must not be compared across arms.
\item \emph{Task design.} The red box is a visual marker that may itself become hard to find for low-resolution models; prior matching of distractors is approximate (the frequency guesser scores above chance); the verification task is balanced by construction but still has a model-specific answer bias. Both tasks use COCO object categories only.
\item \emph{Wording.} Abstention depends on wording (Section~\ref{sec:results}); only two wordings were collected at stage 1, and the verbalized-confidence wording was varied only in the pilot and on a 300-image subset (Section~\ref{sec:verbal}).
\item \emph{Nominal equivalence} is a proxy based on retained contrast; the pre-registered statistic failed, and the gain statistic is a post-hoc replacement.
\item \emph{No human baseline.} We do not know at what visibility people fail on these tasks.
\item \emph{Preprint.} This manuscript has not been peer reviewed.
\end{itemize}

\section{Conclusion}
\label{sec:conclusion}

Placing accuracy, internal and verbalized confidence and abstention on a visibility axis in metres shows, for small open VLMs, a narrow failure band (half-way visibility \VaccFogMin--\VaccFogMax\,m in fog on the identification task) and confidence that stays high well into it, with a blind-spot width of at least a factor of \NLagFactorFog\ in visibility in fog.
An explicit abstain option follows accuracy only when it names the reason, and ImageNet-C fog severities 1--5 lie on the plateau, equivalent to \NomVeqFive--\NomVeqOne\,m.
A verification-style question about the boxed object turned out to measure whether the object is judged visible rather than how well the answer is calibrated, so identification is the better-posed task for confidence studies.
The natural next steps are repeated image draws and more and larger models, a measured or sensor-based depth source, scattering and turbid-water spectra for the underwater arm, a human baseline, and a visibility estimator that triggers abstention before the VLM is asked.

\section*{Data and Code Availability}
The configuration frozen before the main run, the per-question result files (\NRowsTotal\ rows), the analysis scripts and the figure and table generators are available from the authors on request; a public repository will accompany a later version. All datasets and models used are publicly available (COCO val2017, Depth Anything V2, Qwen3-VL, InternVL3.5, SmolVLM2, Gemma-4).

\end{document}